\pdfoutput=1
\documentclass[11pt]{article}

\usepackage[preprint]{acl}
\usepackage{amsmath}
\usepackage{times}
\usepackage{makecell}
\usepackage{threeparttable}
\usepackage{latexsym}
\usepackage{tablefootnote}
\usepackage{graphicx}
\usepackage[labelformat=empty]{subcaption}
\usepackage{listings}
\usepackage[T1]{fontenc}
\usepackage[utf8]{inputenc}

\usepackage{microtype}

\usepackage{inconsolata}

\usepackage{graphicx}
\usepackage{booktabs}
\usepackage{amssymb}
\usepackage{multirow}

\title{\texttt{UNITE-FND}: Reframing Multimodal Fake News Detection through Unimodal Scene Translation}

\author{Arka Mukherjee \\
  KIIT Deemed University \\ Bhubaneswar, India \\
  \texttt{arka.mukherjee078@gmail.com} \\\And
  Shreya Ghosh \\
  IIT Bhubaneswar \\ Bhubaneswar, India \\
  \texttt{shreya@iitbbs.ac.in} \\}

\begin{document}
\maketitle
\begin{abstract}
 Multimodal fake news detection typically demands complex architectures and substantial computational resources, posing deployment challenges in real-world settings. We introduce \texttt{UNITE-FND}\footnote{Codebase: https://anonymous.4open.science/r/UNITE-FND-7118 \\ Full dataset will be made publicly available upon acceptance.}, a novel framework that reframes multimodal fake news detection as a unimodal text classification task. We propose six specialized prompting strategies with Gemini 1.5 Pro, converting visual content into structured textual descriptions, and enabling efficient text-only models to preserve critical visual information. To benchmark our approach, we introduce \textit{Uni-Fakeddit-55k}, a curated dataset family of 55,000 samples each, each processed through our multimodal-to-unimodal translation framework. Experimental results demonstrate that \texttt{UNITE-FND} achieves 92.52\% accuracy in binary classification, surpassing prior multimodal models while reducing computational costs by over 10× (TinyBERT variant: 14.5M parameters vs. 250M+ in SOTA models). Additionally, we propose a comprehensive suite of five novel metrics to evaluate image-to-text conversion quality, ensuring optimal information preservation. Our results demonstrate that structured text-based representations can replace direct multimodal processing with minimal loss of accuracy, making \texttt{UNITE-FND} a practical and scalable alternative for resource-constrained environments.
  
  %Multimodal fake news detection traditionally requires complex architectural components and significant computational resources, limiting its practical application. We present \texttt{UNITE-FND}, a novel framework that democratizes multimodal fake news detection by effectively reducing it to a unimodal text classification task. Our approach leverages Gemini 1.5 Pro's vision-language capabilities through six specialized prompting techniques, converting visual information into structured textual descriptions. This conversion enables the use of efficient text-only models while preserving critical visual information. We introduce Uni-Fakeddit-55k, a carefully curated family of datasets with 55,000 samples, each processed through our six prompting techniques. Experimental results demonstrate that UNITE-FND achieves competitive performance (92.52\% accuracy on binary classification) while significantly reducing computational requirements compared to traditional multimodal approaches. We also propose a comprehensive suite of five metrics for evaluating image-to-text conversion quality, providing insights into information preservation and transfer efficiency. Our framework offers a more accessible approach to multimodal fake news detection, particularly beneficial for resource-constrained environments and real-world applications.
\end{abstract}

\section{Introduction}

The rapid proliferation of multimodal fake news (misleading text combined with manipulated images) has emerged as a major threat to information integrity. Social media platforms such as Instagram, Twitter, and Threads accelerate the spread of deceptive content, making automated detection systems critical for mitigating misinformation~\cite{Aimeur2023FakeNews}. However, existing multimodal fake news detection (FND) methods~\cite{magic2024, gamed2024} often require complex architectures and extensive computational resources, posing significant challenges for real-world deployment.  For instance, GAMED~\cite{gamed2024} and MAGIC~\cite{magic2024} require over 250 million parameters, while self-learning models~\cite{selflearning2024} scale beyond 7 billion parameters. These resource-intensive approaches create a substantial barrier for researchers and organizations lacking access to large-scale infrastructure. Moreover, the complexity of these systems often necessitates expertise in multiple deep learning frameworks and sophisticated deployment strategies.

%Multimodal fake news—combining manipulated images with misleading text—has emerged as a critical threat to information integrity. The proliferation of platforms like Instagram, Twitter, and Threads has accelerated the spread of such content, making automated detection increasingly crucial \cite{Aimeur2023FakeNews}. However, current state-of-the-art approaches to multimodal fake news detection often require substantial computational resources and complex architectures, limiting their practical deployment.

%Recent approaches have achieved impressive accuracy but at a significant computational cost. For instance, GAMED \cite{gamed2024} and MAGIC \cite{magic2024} require over 250 million parameters each, while self-learning approaches \cite{selflearning2024} can exceed 7 billion parameters. These resource requirements create a substantial barrier for smaller organizations and individual researchers who lack access to extensive computational infrastructure. Moreover, the complexity of these systems often necessitates expertise in multiple deep learning frameworks and sophisticated deployment strategies.

\begin{figure*}[t]  % full-width figure at top of page
    \centering
    \includegraphics[width=\textwidth]{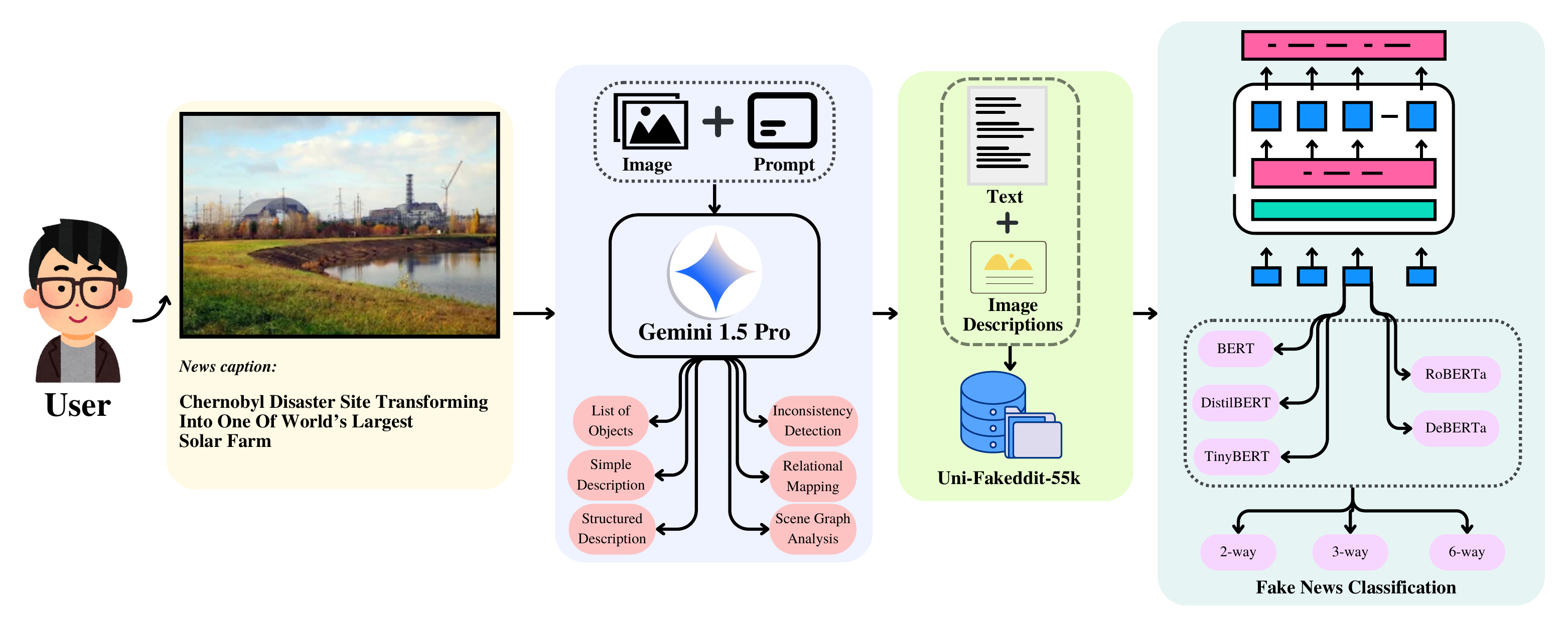} 
    \caption{Overview of the \texttt{UNITE-FND} framework. Our approach transforms multimodal fake news detection into a unimodal task through specialized prompting strategies and efficient text classification.}
    \label{fig:workflow}
\end{figure*}

We argue the prevailing assumption that multimodal fake news detection must rely on heavy multimodal transformers. Instead, we reframe the problem as a unimodal task by leveraging the vision-language capabilities of large multimodal models (VLMs) such as Gemini 1.5 Pro\footnote{https://blog.google/technology/ai/google-gemini-next-generation-model-february-2024/}, GPT-4o\footnote{https://openai.com/index/hello-gpt-4o/}, or Claude 3.5 Sonnet\footnote{https://www.anthropic.com/news/claude-3-5-sonnet}. To the best of our knowledge, this is the first attempt to solve multimodal fake news detection without a dedicated multimodal classifier. 

We introduce \textbf{\texttt{UNITE-FND}} (\textbf{UNI}modal \textbf{T}ranslation \textbf{E}nhanced \textbf{F}ake \textbf{N}ews \textbf{D}etection), a novel framework that eliminates the need for multimodal deep learning architectures. Instead of directly processing multimodal data, \texttt{UNITE-FND} translates images into structured textual representations using six specialized prompting strategies. This enables the use of lightweight text-only classifiers, preserving critical visual information while dramatically reducing computational overhead. Our framework achieves 92.52\% accuracy on 3-way classification using an RTX 4060 laptop GPU, with fine-tuning costs as low as \$1.9, a significant improvement over prior multimodal models requiring cloud-scale resources.

%We argue that multimodality can be simplified to a unimodal task by leveraging the enhanced knowledge base of state-of-the-art large vision-language models (VLMs) such as GPT-4o, Claude 3.5 Sonnet, and Gemini 1.5 Pro. To the best of our knowledge, this is the first attempt at solving multimodal FND with unimodal techniques. We present UNITE-FND (\textbf{UNI}modal \textbf{T}ranslation \textbf{E}nhanced \textbf{F}ake \textbf{N}ews \textbf{D}etection), a novel framework that democratizes multimodal fake news detection by effectively reducing it to a unimodal text classification task. Our approach fundamentally differs from traditional methods by leveraging Gemini 1.5 Pro's vision-language capabilities through six specialized prompting techniques, enabling the use of efficient text-only models while preserving critical visual information. This architectural choice significantly reduces computational requirements—our implementation achieves 91.45\% accuracy on 3-way classification using an RTX 4060 laptop GPU, with model training costs as low as \$2 for competitive performance.

The key contributions of our work include:
\begin{enumerate}
\item \textbf{\texttt{UNITE-FND}}, a framework that transforms multimodal fake news detection into an efficient unimodal classification task, eliminating the need for heavy multimodal models.
    
   % \item UNITE-FND, a novel framework that transforms complex multimodal analysis into an efficient unimodal task\vspace{-0.5em}
   \item We develop \textbf{Uni-Fakeddit-55k}, a dataset containing 55,000 samples derived from the Fakeddit~\cite{nakamura2019r} corpus, processed through six structured prompting techniques to maximize visual information extraction.
    
    %\item \textbf{Uni-Fakeddit-55k}, a family of datasets comprising 55,000 samples from the Fakeddit corpus \cite{nakamura2019r} processed through our prompting techniques for comprehensive visual information extraction\vspace{-0.5em}
    \item A comprehensive suite of five novel metrics for evaluating image-to-text conversion quality, ensuring effective preservation of critical visual cues.\footnote{To support reproducibility and future research, we will publicly release our datasets, prompting templates, and evaluation metrics.
}
    \item We conduct extensive empirical validation, demonstrating that structured text-based representations can replace multimodal models with minimal accuracy loss, enabling efficient inference on consumer-grade hardware (0.2–8.7GB VRAM).
    
    %\item Extensive empirical validation of prompting strategies, demonstrating competitive performance with significantly reduced resource requirements (0.2-8.7GB VRAM)\vspace{-0.5em}
    
\end{enumerate}

Our experimental results demonstrate that \texttt{UNITE-FND} achieves comparable or superior performance to existing approaches while dramatically reducing computational overhead. Using DistilBERT, we achieve 88.75\% accuracy with only 66M parameters and 0.4GB memory usage, compared to recent approaches requiring 250M+ parameters \cite{magic2024}. This efficiency enables deployment on consumer-grade hardware, making sophisticated fake news detection accessible to a broader range of users and organizations.

\section{Related work}

\subsection{Early Fusion-based Approaches}
Early fusion-based approaches established the foundation of multimodal fake news detection, with SpotFake \cite{spotfake2019} pioneering BERT and VGG-19 integration, while EANN \cite{eann2018} introduced event discriminators with VGG and Text-CNN architectures. Basic CNN architectures \cite{basicmm2022} achieved 88\% accuracy on Fakeddit, followed by improvements through SpotFake+ \cite{spotfakeplus2020} (85.6\% on GossipCop) and MVAE \cite{mvae2019} (82.4\% on Weibo). SAFE \cite{safe2020} and CAFE \cite{cafe2021} advanced fusion mechanisms, while HMCAN \cite{hmcan2020} and VERITE \cite{verite2022} introduced attention-based architectures. These early approaches, while groundbreaking, suffered from rigid fusion mechanisms and limited interaction between modalities.

\subsection{Cross-Modal Interaction Frameworks}
Sophisticated approaches focusing on modality interactions aim to solve issues with fusion-based techniques. MIMoE-FND \cite{mimoe2025} achieved 95.6\% accuracy on Weibo-21 through mixture-of-experts architecture. MPFN \cite{mpfn2023} introduced progressive fusion networks with 83.8\% on Weibo. Recently released CroMe \cite{crome2025} achieved 97.4\% on Weibo using tri-transformers through better capture of intra-modality relationships. DAAD \cite{daad2024} and MGCA \cite{mgca2024} further improved modality interactions through dynamic analysis and multi-granularity alignment. They achieve 94.2\% and 91.3\% on Weibo-21 respectively.

\subsection{Large Model Integration}
In recent years, the release of large vision-language models introduced new possibilities,  with FND-CLIP \cite{fndclip2023} achieving 94.2\% on PolitiFact. IMFND \cite{imfnd2024} explores GPT4V and CogVLM, though with limited success (80.1\% on PolitiFact). A self-learning approach \cite{selflearning2024} leverages LLMs for feature extraction without labeled data, achieving 88.88\% on Fakeddit. CMA \cite{cma2024} investigated few-shot learning, though with modest results (79.77\% on PolitiFact). These methods primarily use large models as feature extractors or direct classifiers, whereas \texttt{UNITE-FND} innovatively employs Gemini 1.5 Pro as a modality translator, enabling more effective use of specialized text classification models.

\subsection{Knowledge-Enhanced Detection}
Knowledge-enhanced methods emerged as another direction, with AKA-Fake \cite{akafake2024} employing reinforcement learning and GAMED \cite{gamed2024} introducing multi-expert decoupling (achieving impressive 98.46\% results on the specialized Yang dataset). RaCMC \cite{racmc2024} incorporates multi-granularity constraints with a residual-aware compensation network. These approaches achieve strong performance but require extensive knowledge bases and complex integration mechanisms. Our approach achieves comparable results through Gemini's inherent knowledge, eliminating the need for external knowledge bases.

\subsection{Specialized Approaches}
Specialized approaches have also emerged, including AMPLE \cite{ample2024} with emotion awareness (90\% accuracy on PolitiFact), MMCFND \cite{mmcfnd2024} addressing multilingual challenges across multiple Indic languages (99.6\% on MMIFND), and MAGIC framework \cite{magic2024} using geometric deep learning (98.8\% accuracy on a curated subset of Fakeddit with 3,127 samples).  While these methods excel in specific scenarios, \texttt{UNITE-FND} provides a more generalizable solution through its modality translation.

\section{\texttt{UNITE-FND}}
\texttt{UNITE-FND} redefines multimodal fake news detection by transforming complex multimodal analysis into a streamlined unimodal task. The key innovation lies in leveraging state-of-the-art Large Language Models (LLMs) to convert visual content into structured textual descriptions, enabling efficient text-only classification without sacrificing critical visual information.
%Our proposed methodology introduces a novel approach to multimodal fake news detection by effectively reducing multimodal complexity to unimodal simplicity. The core innovation lies in leveraging state-of-the-art Large Language Models (LLMs) for converting visual information into rich textual descriptions, thereby enabling the use of efficient text-only models for classification.

\subsection{Image-to-Text Conversion Framework}
\begin{figure}[t]
    \centering
    \includegraphics[width=\columnwidth]{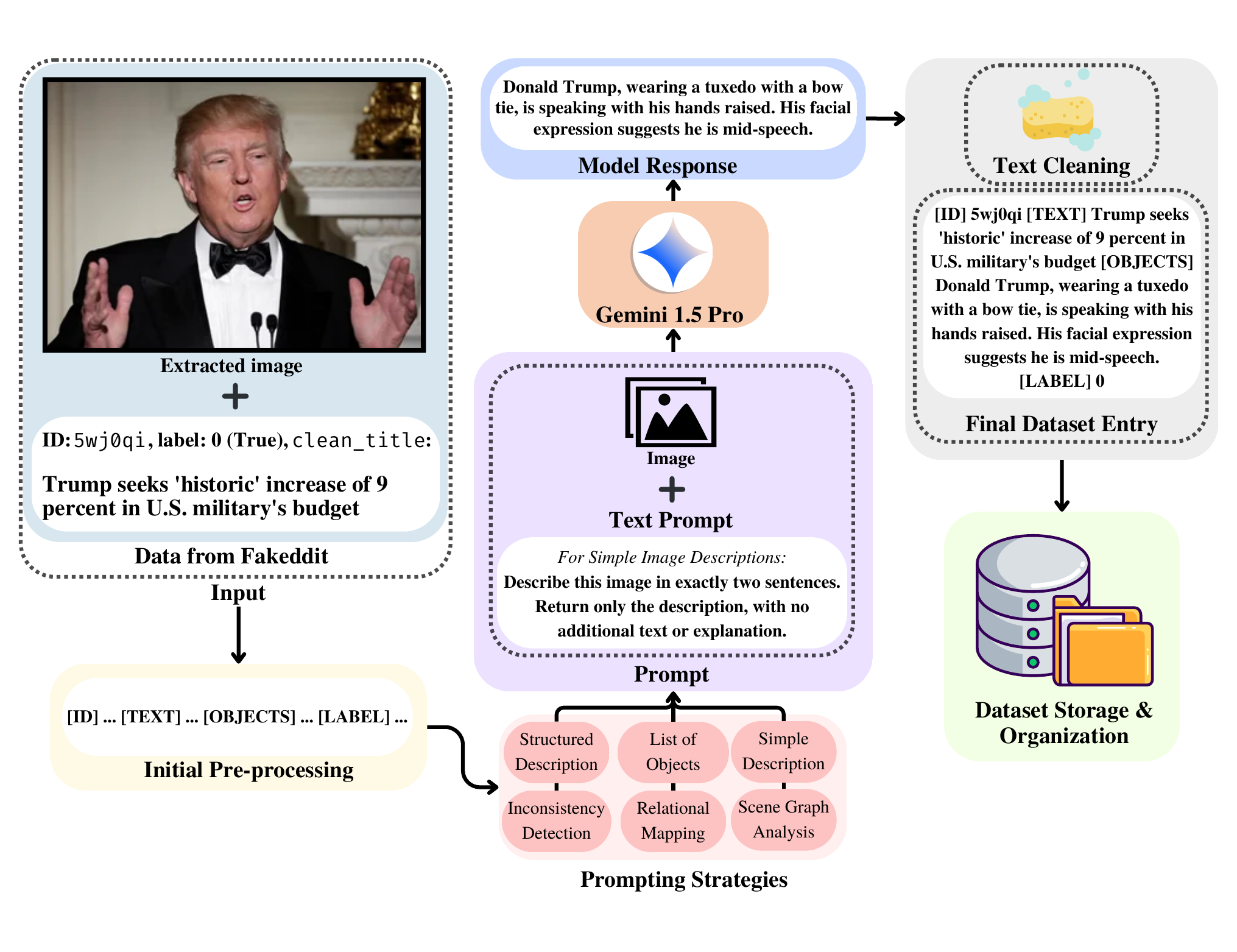}
    \caption{Dataset creation pipeline for Uni-Fakeddit-55k. Each entry from the Fakeddit dataset is processed using six specialized prompting strategies with Gemini 1.5 Pro for image-to-text conversion. The pipeline consists of initial preprocessing, parallel prompting pathways, and structured dataset organization, generating six complementary text-based representations.}
    \label{fig:dataset-workflow}
\end{figure}

At the core of our approach is a robust image-to-text conversion framework powered by Gemini 1.5 Pro, a state-of-the-art multimodal large language model (LLM). We design a structured prompting strategy consisting of six distinct methods, each tailored to extract different aspects of visual information.

%At the heart of our approach is a sophisticated image-to-text conversion framework utilizing Gemini 1.5 Pro, a state-of-the-art multimodal LLM. We developed a comprehensive prompting strategy comprising six distinct methods, each designed to capture different aspects of the visual information:

\subsubsection{Basic Object Identification}
\begin{figure}[ht]
    \centering
    \includegraphics[width=0.47\textwidth]{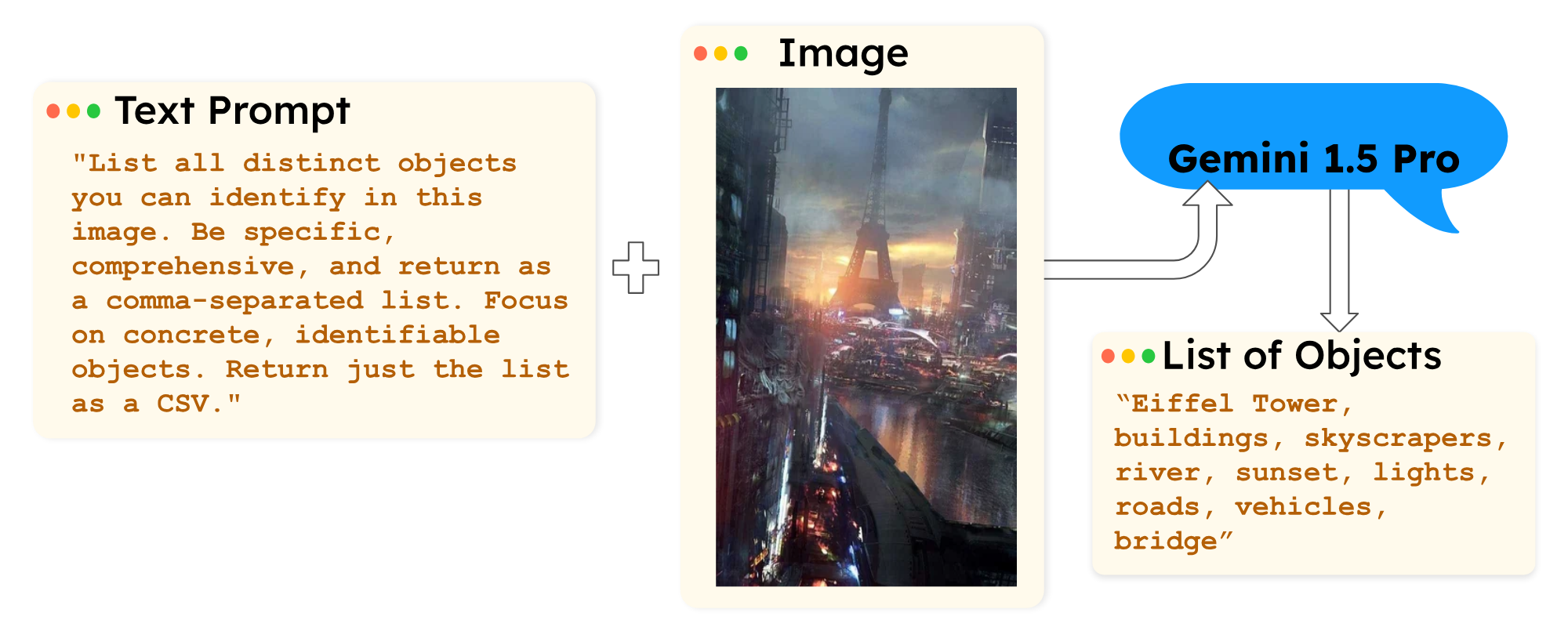}
    \caption{Illustration of the List of Objects prompting approach. The system takes two inputs: (1) a carefully engineered text prompt that requests a comma-separated list of distinct, identifiable objects, and (2) the target image (shown: futuristic cityscape with Eiffel Tower). Gemini 1.5 Pro processes these inputs to generate a structured CSV output containing all major visible objects.}
    \label{fig:list-objects}
\end{figure}

The first method employs a List of Objects approach, utilizing a carefully crafted prompt to generate a comprehensive CSV-formatted inventory of distinct objects within the image. This method ensures the capture of fundamental visual elements while maintaining a structured, machine-processable output format. Figure \ref{fig:list-objects} illustrates our proposed pipeline\footnote{Due to page limitations, examples for each strategy are provided in Appendix \ref{appendix:prompts}}.

\subsubsection{Contextual Description Generation}
To enhance semantic richness, we incorporate two complementary descriptive approaches: Simple Image Description and Structured Image Description. Both methods generate dual-sentence descriptions, with the first sentence capturing observable facts and the second providing contextual interpretation. This dual-layer approach ensures both concrete visual information and implied contextual cues are preserved (Refer Appendix F for examples).

\subsubsection{Spatial Analysis}
The Relational Mapping framework systematically catalogs spatial relationships through a structured JSON schema. Each object is assigned a unique identifier and location descriptor, while relationships between objects are documented with specific interaction types and confidence scores. This approach transforms complex visual-spatial information into a machine-readable format, capturing both direct physical relationships (like "above" or "next to") and interactive associations (such as "facing" or "holding"). The framework prioritizes high-confidence relationships, ensuring reliability in downstream processing while maintaining a clear representation of the image's spatial hierarchy. This method proves particularly valuable for fact-checking claims about spatial arrangements in news photos, such as verifying the authenticity of crowd sizes or the relative positioning of people in event coverage.

\subsubsection{Manipulation Detection Components}
Two specialized components focus on identifying potential image manipulations:

\begin{itemize}
    \item \textbf{Inconsistency Detection:} Performs a comprehensive analysis of visual coherence, examining lighting, perspective, boundaries, and resolution patterns. This component generates structured JSON output detailing potential manipulation indicators with associated confidence scores.
    
    \item \textbf{Scene Graph Analysis:} Creates a detailed representation of the image scene, capturing object relationships, visual quality metrics, and potential manipulation artifacts. This component provides a holistic view of the image's structural integrity.
\end{itemize}
These components are especially effective for scenarios requiring detailed forensic analysis, such as investigating claims of digital tampering in politically sensitive images or verifying the authenticity of emergency situation photographs.

\subsection{Uni-Fakeddit-55k Dataset Creation and Classification Pipeline}

\begin{figure}[ht]
    \centering
    \includegraphics[width=0.5\textwidth]{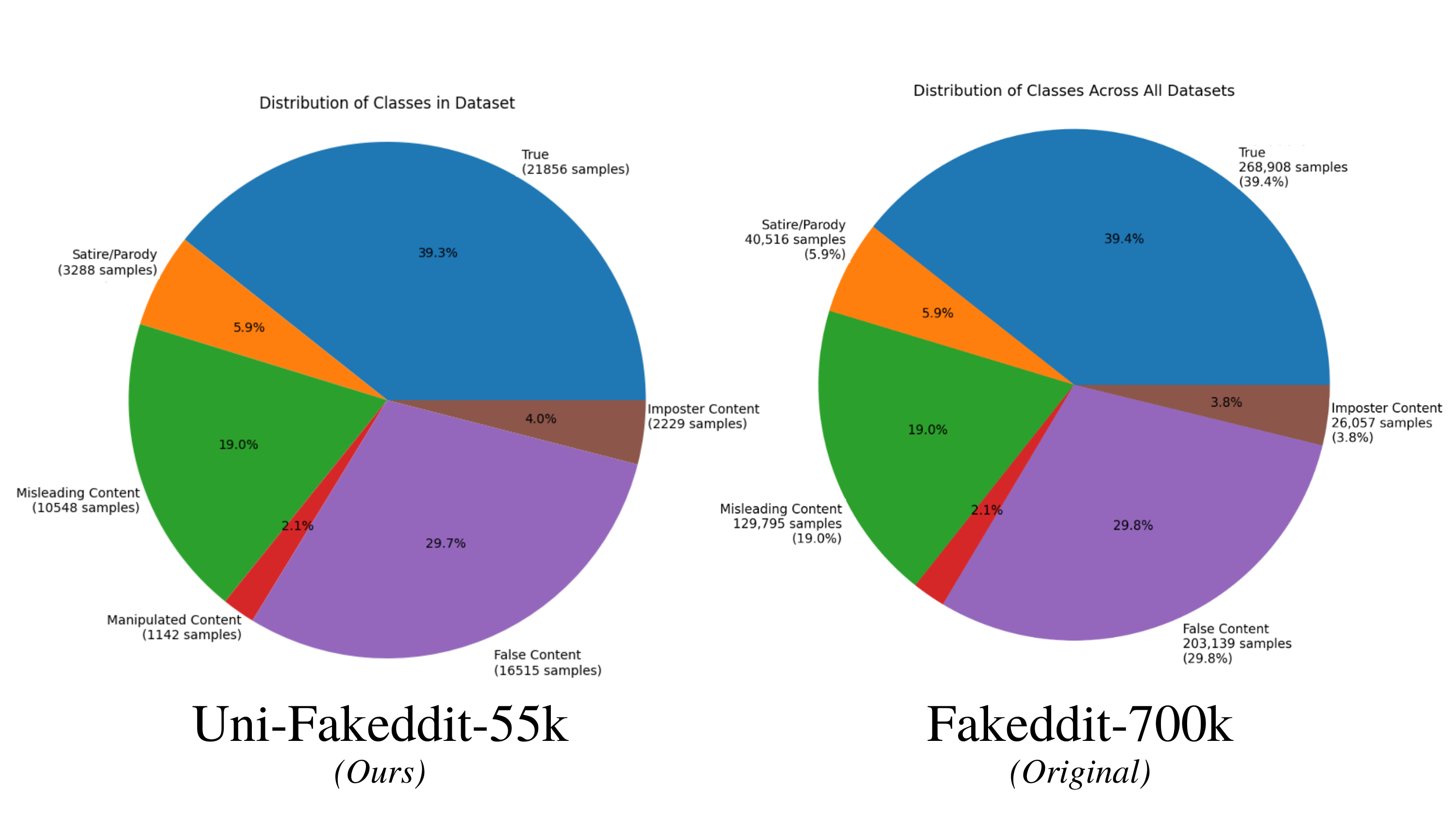}
    \caption{Distribution comparison between our Uni-Fakeddit-55k dataset (left) and the original Fakeddit-700k dataset (right). The pie charts demonstrate that our sampling strategy preserves the relative proportions of different content categories while creating a more manageable dataset size. Both datasets maintain similar class distributions across six categories: True, Satire/Parody, Misleading Content, Manipulated Content, False Content, and Imposter Content, with variations of less than 0.5\% in relative proportions.}
    \label{fig:dataset-comparison}
\end{figure}

Our work introduces a family of six unimodal datasets, collectively called Uni-Fakeddit-55k, derived from the multimodal Fakeddit-700k dataset. Each dataset variant corresponds to one of our six prompting techniques, offering different perspectives on the same underlying content. Figure \ref{fig:dataset-comparison} illustrates the careful sampling strategy employed to maintain class distribution consistency with the original dataset. %We will publicly release all six dataset variants upon acceptance to support reproducibility and further research in unimodal fake news detection. 
More details about the exact prompts and settings used to build the datasets are reported in Appendix \ref{appendix:prompts}.

The textual descriptions generated by our six-component framework undergo several processing steps:

\begin{equation}
T_{final} = f_{merge}(T_{clean}, T_{desc})
\end{equation}

where $T_{clean}$ represents the cleaned news title, $T_{desc}$ represents the concatenated image descriptions, and $f_{merge}$ is our custom text merger function.

The processed text is then passed through a selection of text-only models: BERT, DistilBERT, TinyBERT, RoBERTa-Large, and DeBERTa-V3-Large.

Each model is fine-tuned on our processed dataset using the following objective:

\begin{equation}
\mathcal{L} = -\sum_{i=1}^{N} y_i \log(\hat{y_i}) + \lambda \|\theta\|_2
\end{equation}

where $y_i$ represents the true label, $\hat{y_i}$ is the model prediction, and $\lambda$ controls the L2 regularization strength.

Using this pipeline, we effectively transform the complex task of multimodal fake news detection into a more tractable text-only classification problem while preserving critical visual information through our sophisticated image-to-text conversion framework.

\subsection{\texttt{UNITE-FND} Model Architecture: Optimized for Resource-Constrained Environments}

While recent approaches have achieved impressive accuracy through increasingly complex architectures, \texttt{UNITE-FND} prioritizes efficiency without significant performance trade-offs. As shown in Table~\ref{tab:model-comparison}, current state-of-the-art approaches like MAGIC~\cite{magic2024} and GAMED~\cite{gamed2024} require over 250 million parameters, while self-learning approaches exceed 7 billion parameters. In contrast, our most efficient implementation using TinyBERT requires only 14.5 million parameters while achieving 87.4\% accuracy on Uni-Fakeddit-55k. Even our largest model variant using DeBERTa-large (400M parameters) requires significantly fewer parameters than certain contemporary approaches while achieving a competitive accuracy of 92.5\%.

This efficiency translates directly into practical advantages: (a) Our models require as little as 0.2GB VRAM (TinyBERT) and scale up to just 8.7GB (DeBERTa), enabling deployment on consumer-grade hardware without the need for specialized infrastructure. (b) The cost-effectiveness of our approach is equally notable, with full model fine-tuning achievable for as little as \$2 using cloud-based compute resources. (c) Additionally, the flexibility of \texttt{UNITE-FND} allows organizations to select model variants that best align with their accuracy-efficiency trade-offs, making it adaptable to diverse operational constraints. These characteristics position \texttt{UNITE-FND} as a highly scalable solution for resource-constrained environments while maintaining performance competitive with significantly more complex architectures.

%This efficiency translates directly to practical benefits:
%\begin{itemize}
   % \item \textbf{Reduced Memory Usage:} Our models require between 0.2GB (TinyBERT) and 8.7GB (DeBERTa) VRAM, making them deployable on consumer-grade hardware
   % \item \textbf{Cost-Effective Training:} Complete model fine-tuning can be accomplished for as little as \$2 of compute resources using cloud infrastructure
   % \item \textbf{Flexible Deployment:} Organizations can choose model variants based on their specific accuracy-efficiency trade-off requirements
%\end{itemize}

%These characteristics make UNITE-FND particularly suitable for resource-constrained environments while maintaining competitive performance with more complex architectures.

\section{Evaluation Metrics}
To comprehensively evaluate the quality and effectiveness of our image-to-text conversion framework, we propose a novel suite of five complementary metrics, each designed to capture different aspects of information preservation and transfer quality. These metrics are then combined into a unified Composite Information Quality Score (CIQS).

\subsection{Image Preservation Rate (IPR)}
The Image Preservation Rate quantifies how effectively the textual description retains the essential information present in the original image. We improve upon traditional correlation-based approaches by introducing a non-linear transformation that better captures the relationship between feature spaces. Let $I$ be the image features and $T$ be the text embeddings:

\begin{equation}
IPR(I,T) = 1 - e^{-5s}
\end{equation}

where $s$ represents the scaled cosine similarity between normalized projections $P_I$ and $P_T$ in a common space of dimension $d = min(dim(I), dim(T))$:

\begin{equation}
s = \frac{cos(P_I, P_T) + 1}{2}
\end{equation}

The projections are obtained through Xavier-initialized linear transformations to ensure stable feature mapping across modalities.

\subsection{Semantic Coverage Score (SCS)}

SCS evaluates the comprehensiveness of the generated description, focusing on object-centric content. For object list techniques, we employ a multi-criteria evaluation:

\begin{equation}
SCS(T) = w_l \cdot L + w_s \cdot S + w_c \cdot C
\end{equation}

\noindent where \( L = \min(|O|/10, 1.0) \) represents the normalized object count, \( S \) quantifies specificity by penalizing generic terms, and \( C \) measures completeness based on the presence of multi-word descriptions. The weighting factors are set as \( w_l = 0.3 \), \( w_s = 0.4 \), and \( w_c = 0.3 \).

%SCS evaluates the comprehensiveness of the generated description with particular attention to object-centric content. For object list techniques, we employ a multi-criteria evaluation:

%\begin{equation}
%SCS(T) = w_l \cdot L + w_s \cdot S + w_c \cdot C
%\end{equation}

%where:
%\begin{itemize}
%\item $L = min(|O|/10, 1.0)$ represents normalized object count\vspace{-0.5em}
%\item $S$ measures specificity through absence of generic terms\vspace{-0.5em}
%\item $C$ evaluates completeness via multi-word descriptions\vspace{-0.5em}
%\item $w_l, w_s, w_c$ are weights (0.3, 0.4, 0.3 respectively)\vspace{-0.5em}
%\end{itemize}

\subsection{Information Specificity Score (ISS)}
ISS measures the semantic depth and specificity of the generated description:

\begin{equation}
ISS(T) = \frac{1}{|W|} \sum_{w\in W} \frac{D(w)}{D_{max}}
\end{equation}

\noindent where $W$ is the set of content words, $D(w)$ represents the WordNet depth of word $w$, and $D_{max}$ is the maximum WordNet depth (approximately 20).

\subsection{Structural Information Retention (SIR)}
SIR evaluates the preservation of structural relationships, with distinct formulations for different technique types:

\noindent (1) For graph-based techniques:
\begin{equation}
SIR(G) = \frac{1}{4}(N_s + E_d + R_d + C_s)
\end{equation}

\noindent where:
%\begin{itemize}
$N_s = min(|V|/10, 1)$ represents the node score, $E_d$ is the edge density, $R_d = min(|R|/5, 1)$ measures relationship diversity and $C_s = \frac{conf_V + conf_E}{2}$ indicates confidence score.
\vspace{0.4cm}

\noindent (2) For text-based techniques:
\begin{equation}
SIR(T) = \frac{|S|}{5}
\end{equation}
where $S$ is the set of sentences.

\subsection{Modality Transfer Efficiency (MTE)}
MTE assesses both the efficiency of information transfer and preservation of feature complexity:

\begin{equation}
MTE(I,T) = 0.7 \cdot S + 0.3 \cdot C_r
\end{equation}

\noindent where $S$ represents cosine similarity between normalized projections, $C_r = \frac{min(\sigma_I, \sigma_T)}{max(\sigma_I, \sigma_T)}$ measures complexity ratio, $\sigma_I, \sigma_T$ are standard deviations of projected features.

\subsection{Composite Information Quality Score (CIQS)}
Finally, we combine all metrics into a single comprehensive score:

\begin{equation}
CIQS = (IPR \cdot SCS \cdot ISS \cdot SIR \cdot MTE)^{\frac{1}{5}}
\end{equation}

\noindent This geometric mean ensures a balanced contribution from all components while penalizing poor performance in any single metric. The CIQS provides a holistic assessment of the image-to-text conversion quality, considering preservation, coverage, specificity, structure, and transfer efficiency.

\section{Results}

\subsection{Quality of Image-to-Text Conversion}
We first evaluate the quality of our image-to-text conversion strategies using our proposed metrics suite (Table \ref{tab:prompting_metrics}), where each metric is normalized to a [0,1] range except for SIR which can exceed 1 due to its structural complexity measure. All strategies demonstrate strong Information Preservation Rate (IPR > 0.91), indicating effective retention of core visual information. Scene Graph Analysis and Inconsistency Detection achieve particularly high Semantic Coverage Scores (SCS > 0.96), suggesting comprehensive capture of image content. While List of Objects shows the highest Information Specificity Score (ISS = 0.4635), Scene Graph Analysis excels in Structural Information Retention (SIR = 2.1231), reflecting its superior ability to preserve relational information. Modality Transfer Efficiency (MTE) remains consistent across all strategies (~0.65), indicating stable information transfer between modalities.

\begin{table*}[t]
    \centering
    \begin{tabular}{lccccccc}
    \hline
    \textbf{Prompting Strategy} & \textbf{IPR} & \textbf{SCS} & \textbf{ISS} & \textbf{SIR} & \textbf{MTE} & \textbf{CIQS} \\
    \hline
    List of Objects & 0.9176 & 0.7417 & 0.4635 & 0.2030 & 0.6498 & 0.5268 \\
    Simple Image Description & 0.9176 & 0.7690 & 0.3925 & 0.4049 & 0.6497 & 0.5910 \\
    Structured Image Description & 0.9175 & 0.8148 & 0.3727 & 0.3805 & 0.6498 & 0.5827 \\
    Relational Mapping & 0.9176 & 0.9243 & 0.4102 & 0.2055 & 0.6497 & 0.5396 \\
    Inconsistency Detection & 0.9175 & 0.9780 & 0.2802 & 0.9443 & 0.6497 & 0.6742 \\
    Scene Graph Analysis & 0.9176 & 0.9668 & 0.3677 & 2.1231 & 0.6497 & 0.8484 \\
    \hline
    \end{tabular}
    \caption{Evaluation Metrics Across Different Prompting Strategies on Uni-Fakeddit-55k}
    \label{tab:prompting_metrics}
\end{table*}

\subsection{Classification Performance}
\subsubsection{Comparison with Vision-Language Models }
\texttt{UNITE-FND} framework achieves state-of-the-art performance on the Uni-Fakeddit-55k dataset, significantly outperforming existing vision-language models (Table \ref{tab:fakeddit_accuracy}). The framework achieves 92.52\% accuracy in binary classification, substantially surpassing recent models like Llama-3.2-11B-Vision (63.92\%) and Video-LLaVA-7B (59.34\%). This marked improvement demonstrates the effectiveness of our modality translation approach over direct multimodal processing.

\begin{table}[ht]
    \centering
    \begin{tabular}{p{4.0 cm}c}
    \hline
    \textbf{Model} & \textbf{Accuracy (\%)} \\
    \hline
    \texttt{LanguageBind/}\\
    \texttt{Video-LLaVA-7B-hf} & 59.34 \\
    \texttt{Salesforce/}\\
    \texttt{instructblip-vicuna-7b} & 59.35 \\
    \texttt{meta-llama/}
    \\{\texttt{Llama-3.2-11B-Vision}} & 63.92 \\
    \midrule
    \textbf{\texttt{UNITE-FND} (Ours)} & \textbf{92.52} \\
    \hline
    \end{tabular}
    \caption{Binary Classification Accuracy on Uni-Fakeddit-55k Dataset}
    \label{tab:fakeddit_accuracy}
\end{table}

\subsubsection{Comparison with Existing Fake News Detection Model}
When compared to existing multimodal fake news detection systems (Table \ref{tab:performance-metrics}), \texttt{UNITE-FND} demonstrates substantial improvements across all metrics. Our framework achieves gains of:
\begin{itemize}%\setlength{\parskip}{0pt}\setlength{\itemsep}{0pt}
\item \textbf{Accuracy:}  +3.64\% (↑ over Self-learning \cite{selflearning2024})
  \item  \textbf{Precision:}  +6.16\% (↑ over SpotFake+ \cite{spotfakeplus2020}) 
   \item \textbf{Recall:}  +7.23\% (↑ over CAFE \cite{cafe2021})
   \item \textbf{F1-score:}  +6.63\% (↑ over Self-learning \cite{selflearning2024})

%\item +3.64 percentage points in accuracy over the previous best (Self-learning \cite{selflearning2024})
%\item +6.16 percentage points in precision over SpotFake+ \cite{spotfakeplus2020}
%\item +7.23 percentage points in recall over CAFE \cite{cafe2021}
%\item +6.63 percentage points in F1-score over Self-learning \cite{selflearning2024}
\end{itemize}

These results demonstrate that our approach of converting multimodal fake news detection into a unimodal problem through sophisticated image-to-text conversion not only simplifies the architecture but also leads to superior performance. The consistent improvements across all metrics suggest that our framework better captures the nuanced relationships between visual and textual content in fake news detection.

\begin{table}[t]
    \centering
    \begin{tabular}{p{2.5 cm}lccccc}
        \toprule
        \textbf{Model} & \textbf{Acc} & \textbf{P} & \textbf{R} & \textbf{F1} \\
        \midrule
        MVAE \cite{mvae2019} & 70.24 & 76.53 & 74.75 & 75.63 \\
        EANN \cite{eann2018} & 72.27 & 78.43 & 63.40 & 70.12 \\
        SpotFake \cite{spotfake2019} & 77.29 & 71.63 & 70.77 & 71.20 \\
        HMCAN \cite{hmcan2020} & 82.89 & 84.03 & 84.04 & 84.03 \\
        SpotFake+ \cite{spotfakeplus2020} & 83.08 & 86.38 & 84.87 & 85.62 \\
        CAFE \cite{cafe2021} & 84.14 & 85.39 & 85.27 & 85.32 \\
        VERITE \cite{verite2022} & 84.72 & 85.34 & 84.37 & 84.85 \\
        Self-learning \cite{selflearning2024} & 88.88 & 86.40 & 85.40 & 85.90 \\
        \midrule
        \textbf{\texttt{UNITE-FND} (Ours)} & \textbf{92.52} & \textbf{92.56} & \textbf{92.50} & \textbf{92.53} \\
        \bottomrule
    \end{tabular}
    \begin{tablenotes}
      
      \item Note: All metrics are percentages. Experiments conducted on a 10\% subset of the Fakeddit dataset for fair comparison.
    \end{tablenotes}
    \caption{Performance Comparison of Fake News Detection Models on Fakeddit (10\% subset)}
    \label{tab:performance-metrics}
\end{table}
While our results demonstrate strong performance, it is essential to consider variations in evaluation metrics when comparing across different dataset splits. Prior work~\cite{gamed2024} shows that models such as EANN and MVAE achieve substantially higher accuracy on the full Fakeddit dataset (87.5\% and 88.75\%, respectively) compared to their performance on a 10\% subset (72.27\% and 70.24\%). More recent advancements further push performance boundaries, with GAMED reaching 93.93\%, followed by CLIP+LLaVA at 92.54\% and BMR at 91.65\%. These trends suggest that model performance benefits significantly from larger training sets. Although direct comparisons across different dataset configurations must be interpreted cautiously, it is important to note that \texttt{UNITE-FND} achieves 92.52\% accuracy using only a 10\% subset of Fakeddit. Despite being trained on significantly less data, our approach remains highly competitive with state-of-the-art multimodal models evaluated on the full dataset. This underscores the effectiveness of structured image-to-text conversion in preserving critical visual information while reducing computational complexity.

%However, it's important to note that performance metrics can vary significantly when evaluating on the complete Fakeddit dataset. Recent work by \cite{gamed2024} demonstrates that methods like EANN and MVAE achieve substantially higher performance on the full corpus (87.5\% and 88.75\% accuracy respectively) compared to their results on the 10\% subset (72.27\% and 70.24\%).
%More recent approaches on the full Fakeddit dataset have pushed performance boundaries further, with GAMED achieving 93.93\% accuracy, followed by CLIP+LLaVA at 92.54\% and BMR at 91.65\%. These results suggest that model performance benefits significantly from the larger training set. While direct comparisons across different dataset splits require careful consideration, UNITE-FND's performance (92.52\% accuracy) remains competitive with these state-of-the-art approaches, particularly notable given our framework's architectural simplicity compared to more complex multimodal architectures.

\subsection{Observations on Proposed Evaluation Metrics}

Analysis of our evaluation metrics (Table~\ref{tab:prompting_metrics}) reveals distinct patterns that offer key insights into the effectiveness of different prompting strategies. All techniques exhibit consistently high Information Preservation Rates (IPR $\approx$ 0.917), which explains the substantial improvements observed over text-only baselines.

\noindent \underline{Observation 1.} The Semantic Coverage Score (SCS) highlights a clear distinction between simpler and more sophisticated approaches. JSON-based methods such as Scene Graph (0.9668) and Inconsistency Detection (0.9780) achieve significantly higher coverage than simpler strategies like List of Objects (0.7417). However, this increased coverage does not translate into proportional classification gains, suggesting that transformer models may not fully exploit the additional structured information.

\noindent \underline{Observation 2.} A notable inverse relationship emerges between technique sophistication and Information Specificity Scores (ISS). Simpler methods, such as List of Objects (0.4635), outperform more complex strategies like Scene Graph (0.3677). This pattern suggests that while structured representations capture extensive visual information, they may introduce noise that hinders model interpretability.

\noindent \underline{Observation 3.} The Structural Information Retention (SIR) metric shows considerable variation, with Scene Graph Analysis achieving the highest score (2.1231) due to its detailed encoding of spatial relationships. Interestingly, Relational Mapping performs worse (0.2055) than even basic descriptions, likely because it emphasizes object interactions rather than spatial structure.

\noindent \underline{Observation 4.} The Composite Information Quality Score (CIQS) presents an unexpected discrepancy: Scene Graph Analysis achieves the highest CIQS (0.8484) but ranks second in classification performance, whereas Structured Image Description, which ranks fourth in CIQS (0.5827), consistently outperforms other techniques across transformer architectures. This suggests that \textit{maximizing information capture alone is insufficient}; rather, effective fake news detection requires balancing information complexity with model interpretability and processing efficiency.

These findings highlight the importance of our proposed evaluation metrics that not only capture rich multimodal information but also align with model interpretability and processing capabilities. By providing deeper insights into the relationship between structured textual representations and classification performance, our metrics offer a valuable framework for optimizing multimodal-to-unimodal transformations in misinformation detection.

\section{Conclusion and Future Works}
We present \texttt{UNITE-FND}, a novel framework that reframes multimodal fake news detection as a unimodal task through structured image-to-text conversion. Our approach significantly reduces computational overhead while maintaining competitive accuracy, enabling deployment on consumer-grade hardware. Experimental results demonstrate that structured textual representations effectively replace direct multimodal processing with minimal performance loss, highlighting the potential of language models in preserving critical visual information. %Despite its efficiency, transitioning to a unimodal paradigm introduces challenges in information preservation, particularly in capturing implicit visual cues. 
Future work can explore adaptive prompt optimization, multilingual adaptation, and temporal modeling to enhance robustness. Additionally, improving explainability and mitigating adversarial vulnerabilities will be crucial for responsible deployment. By balancing efficiency, accessibility, and interpretability, \texttt{UNITE-FND} lays the foundation for scalable, real-world misinformation detection.

\section*{Limitations}
While \texttt{UNITE-FND} shows promising results, there are few limitations. First, our approach's reliance on VLMs (specifically Gemini 1.5 Pro) for image-to-text conversion introduces a potential bottleneck in processing speed. Second, the quality of text generation can vary based on the VLM's understanding of complex visual scenarios. Third, while our method significantly reduces computational requirements for deployment, it still requires access to the Gemini API, which may present cost considerations for large-scale applications.

\section*{Ethical Considerations}

Our work utilizes the Fakeddit dataset, which contains Reddit posts that have been carefully curated and pre-processed by its creators to address privacy and ethical concerns. Given the dataset's thorough preparation, which includes removal of personally identifiable information and offensive content, we maintained its existing privacy safeguards without additional processing to preserve data integrity and statistical properties. Our image-to-text conversion strategies are designed to extract only objective visual elements, ensuring that generated descriptions exclude references to specific individuals or private information. While our approach enhances accessibility and efficiency in fake news detection, we recognize the potential risks associated with its misuse, including the possibility of censorship or content manipulation. We advocate for responsible and
transparent use, respecting individual privacy and freedom
of expression, with clear communication about its deployment and the option for users to opt-out. We acknowledge
the possibility for potential false positives and false negatives, and we suggest continuous research, development, and
stakeholder feedback for system refinement. We declare no competing interests. The research was conducted independently, and the framework
was developed for academic and public benefit aiming to
better understand and fight misinformation online.

\bibliography{acl_latex}
\clearpage
\appendix

\noindent{\LARGE \textbf{Appendices}}  % Large, bold title without section numbering

\vspace{0.5cm}
% \label{sec:appendix}

This supplementary material presents additional details on the following aspects:  
\begin{itemize}
    \setlength{\itemsep}{2pt}
    \item \textbf{Appendix A:} Literaure Review (in details)
    \item \textbf{Appendix B:} Model Architecture and Performance Studies
    \item \textbf{Appendix C:} Ablation Studies
    \item \textbf{Appendix D:} Training Costs 
    \item \textbf{Appendix E:} Gemini 1.5 Pro - Why?
     \item \textbf{Appendix F:} Prompting Techniques and Examples 
     \item \textbf{Appendix G:} Training Hyperparameters 
     \item \textbf{Appendix H:} Vision-Language Model Testing Configuration 
\end{itemize}

\section{Literature Review}
\label{appendix:literature}

\begin{table*}
%\small
  \label{tab:fnd-comparison}
  \begin{tabular}{p{5.0 cm}p{5.58cm}c}
    \toprule
    \textbf{Method} & \textbf{Key Components} & \textbf{Best Performance (Dataset)}\\
    \midrule
    SpotFake+ \cite{spotfakeplus2020} & BERT + VGG-19 & 85.6\% (GossipCop)\\
    SAFE \cite{safe2020} & Similarity-aware fusion & 87.4\% (PolitiFact)\\
    MM-CNN \cite{basicmm2022} & Multimodal CNNs & 87.0\% (Fakeddit)\\
    MPFN \cite{mpfn2023} & Progressive fusion & 83.8\% (Weibo)\\
    FND-CLIP \cite{fndclip2023} & CLIP + feature concat & 94.2\% (PolitiFact)\\
    MAGIC \cite{magic2024} & Graph Neural Networks & 98.8\% (Fakeddit)*\\
    IMFND \cite{imfnd2024} & GPT4V + Zero-shot & 80.1\% (PolitiFact)\\
    MMCFND \cite{mmcfnd2024} & Multilingual encoders & 99.6\% (MMIFND)\\
    GAMED \cite{gamed2024} & Modal decoupling + Expert networks & 93.93\% (Fakeddit)\\
    AKA-Fake \cite{akafake2024} & Knowledge graphs + RL & 91.9\% (PolitiFact)\\
    CroMe \cite{crome2025} & Tri-Transformer + BLIP2 & 97.4\% (Weibo)\\
    MIMoE-FND \cite{mimoe2025} & Mixture-of-Experts & 95.6\% (Weibo-21)\\
    \midrule
    \textbf{UNITE-FND (Ours)} & \textbf{Multi-strategy Vision Translation + DeBERTa} & \textbf{92.52\% (Fakeddit)}\\
    \bottomrule
    \vspace{1mm}
    \\[-2.5ex]
    \multicolumn{3}{l}{*Using a carefully curated set of just 3,127 samples from Fakeddit}
  \end{tabular}
  \caption{Comparison of Recent Multimodal Fake News Detection Approaches}
\end{table*}

This appendix provides a comprehensive review of multimodal fake news detection approaches, organized by their methodological focus and chronological development.
\subsection{Early Fusion-based Approaches (2019-2022)}
Early work in multimodal fake news detection established foundational architectures for combining visual and textual information:
\begin{itemize}\setlength{\parskip}{0pt}\setlength{\itemsep}{0pt}
\item SpotFake \cite{spotfake2019} pioneered multimodal fusion using BERT and VGG-19, achieving 89.23\% accuracy on Weibo
\item MVAE \cite{mvae2019} introduced a bimodal variational autoencoder approach, reaching 82.4\% accuracy on Weibo
\item SpotFake+ \cite{spotfakeplus2020} enhanced the architecture with XLNet, achieving 85.6\% on GossipCop
\item SAFE \cite{safe2020} introduced similarity-aware fusion, reaching 87.4% on PolitiFact
\item MM-CNN \cite{basicmm2022} demonstrated that basic CNN architectures could achieve 87.0\% accuracy on Fakeddit
\end{itemize}
\subsection{Advanced Architectures (2023-2024)}
Recent approaches have introduced more sophisticated architectural innovations:
\begin{itemize}\setlength{\parskip}{0pt}\setlength{\itemsep}{0pt}
\item MPFN \cite{mpfn2023} explored progressive fusion to capture both shallow and deep features (83.8\% on Weibo)
\item FND-CLIP \cite{fndclip2023} leveraged CLIP's pre-trained representations (94.2\% on PolitiFact)
\item MAGIC \cite{magic2024} introduced graph neural networks, achieving 98.8\% on a curated Fakeddit subset
\item GAMED \cite{gamed2024} employed modal decoupling with expert networks (93.93\% on Fakeddit)
\item RaCMC \cite{racmc2024} introduced residual-aware compensation networks (92.2\% on Weibo-21)
\end{itemize}
\subsection{Knowledge-Enhanced Methods}
Several approaches have incorporated external knowledge:
\begin{itemize}\setlength{\parskip}{0pt}\setlength{\itemsep}{0pt}
\item AKA-Fake \cite{akafake2024} used reinforcement learning for knowledge graph construction (91.9\% on PolitiFact)
\item AMPLE \cite{ample2024} integrated emotion-aware analysis (90\% on PolitiFact)
\item DAAD \cite{daad2024} introduced dynamic analysis mechanisms (94.2\% on Weibo-21)
\end{itemize}
\subsection{Large Model Integration}
Recent work has explored the potential of large vision-language models:
\begin{itemize}\setlength{\parskip}{0pt}\setlength{\itemsep}{0pt}
\item IMFND \cite{imfnd2024} investigated zero-shot capabilities of GPT4V (80.1\% on PolitiFact)
\item CroMe \cite{crome2025} combined BLIP2 with tri-transformers (97.4\% on Weibo)
\item MIMoE-FND \cite{mimoe2025} introduced mixture-of-experts architecture (95.6\% on Weibo-21)
\end{itemize}
\subsection{Specialized Applications}
Several approaches have focused on specific challenges:
\begin{itemize}\setlength{\parskip}{0pt}\setlength{\itemsep}{0pt}
\item MMCFND \cite{mmcfnd2024} addressed multilingual detection, achieving 99.6\% on MMIFND
\item Self-Learning FND \cite{selflearning2024} explored unlabeled data utilization (88.88\% on Fakeddit)
\item MGCA \cite{mgca2024} introduced multi-granularity clue alignment (91.3\% on Weibo-21)
\end{itemize}
\subsection{Dataset Usage Patterns}
Analysis of the literature reveals several commonly used datasets:
\begin{itemize}\setlength{\parskip}{0pt}\setlength{\itemsep}{0pt}
\item Weibo and Weibo-21: Popular for Chinese content evaluation
\item PolitiFact and GossipCop: Standard benchmarks for English news
\item Fakeddit: Largest dataset, often used for a comprehensive evaluation
\item Specialized datasets: MMIFND (multilingual), Yang (domain-specific)
\end{itemize}
This comprehensive review demonstrates the field's evolution from simple fusion approaches to sophisticated architectures incorporating large models and specialized techniques. Our \texttt{UNITE-FND} framework builds upon these advances while introducing a novel approach to modality translation that achieves competitive performance with reduced computational requirements.

\section{Model Architecture and Performance Studies}
\label{appendix:architecture}

\begin{table*}[t]
    \centering
    
    \begin{tabular}{p{4.5 cm}llrrr}
        \toprule
        \textbf{Paper} & \textbf{Model} & \textbf{Parameters} & \textbf{Dataset} & \textbf{Accuracy} \\
        \midrule
        \multirow{4}{*}{\parbox{4.2cm}{Multimodal FND\\\cite{basicmm2022}}}
            & Unimodal CNN & 15.2M & \multirow{4}{*}{Fakeddit} & 74.0\% \\
            & Unimodal BiLSTM & 15.3M & & 75.0\% \\
            & Unimodal BERT & 110M & & 76.0\% \\
            & Multimodal CNN & 22.4M & & 88.0\% \\
        \midrule
        GAMED \cite{gamed2024} & GAMED & 258M & \makecell[r]{Fakeddit\\Yang} & \makecell[r]{93.9\%\\98.5\%} \\
        \midrule
        MAGIC \cite{magic2024} & MAGIC & 252M & \makecell[r]{Fakeddit\\MChinese} & \makecell[r]{98.8\%\\86.3\%} \\
        \midrule
        \makecell[l]{Self-Learning FND\\\cite{selflearning2024}} & N/A & 7.2B-13.2B & Fakeddit & 88.9\% \\
        \midrule
        \multirow{4}{*}{\textbf{\texttt{UNITE-FND} (Ours)}} 
            & TinyBERT & 14.5M & \multirow{4}{*}{\makecell[r]{Uni-Fakeddit\\-55k}} & 87.4\% \\
            & DistilBERT & 66M & & 88.8\% \\
            & BERT-base & 110M & & 89.2\% \\
            & RoBERTa-large & 355M & & 91.5\% \\
            & DeBERTa-large & 400M & & 92.5\% \\
        \bottomrule
    \end{tabular}
    \begin{tablenotes}
      \item Note: Parameter counts are rounded to the nearest million (M) or billion (B). Accuracy reported for Fakeddit is on 2-way classification unless otherwise specified.
    \end{tablenotes}
    \caption{Comparison of Model Architectures and Performance Metrics}
    \label{tab:model-comparison}
\end{table*}

While the main paper focuses on \texttt{UNITE-FND}'s key contributions and primary results, this appendix provides a detailed comparison of model architectures and their respective performance metrics. Table~\ref{tab:model-comparison} presents a comprehensive overview of recent approaches in multimodal fake news detection, highlighting the relationship between model complexity and detection accuracy.

Current state-of-the-art approaches demonstrate a clear trend toward increasingly complex architectures. MAGIC~\cite{magic2024} and GAMED~\cite{gamed2024} employ sophisticated neural networks with over 250 million parameters each, while self-learning approaches~\cite{selflearning2024} utilize large language models ranging from 7.2 to 13.2 billion parameters. These approaches achieve impressive accuracy but at substantial computational cost.

\texttt{UNITE-FND}, in contrast, demonstrates that efficient architectures can achieve competitive performance through effective modality translation. Our implementation spans a range of model sizes:
\begin{itemize}\setlength{\parskip}{0pt}\setlength{\itemsep}{0pt}
\item TinyBERT (14.5M parameters) achieves 87.4\% accuracy, matching the performance of more complex multimodal architectures while using fewer parameters than even basic unimodal CNNs from earlier approaches~\cite{basicmm2022}
\item DistilBERT (66M parameters) and BERT-base (110M parameters) demonstrate that moderate-sized models can achieve strong performance (88.8\% and 89.2\% respectively)
\item Our larger models, RoBERTa-large (355M parameters) and DeBERTa-large (400M parameters), achieve competitive accuracy (91.5\% and 92.5\%) while still maintaining significantly lower parameter counts than contemporary approaches
\end{itemize}

This analysis supports our main paper's argument that effective modality translation can enable simpler, more efficient architectures to achieve competitive performance in fake news detection. The consistent performance across model sizes suggests that our approach's effectiveness stems from the quality of the image-to-text conversion rather than raw model capacity.

\section{Ablation Studies}
\label{appendix:ablations}

Detailed analysis of model performance across different architectures (Table \ref{tab:classification-results}) reveals several significant patterns and insights. RoBERTa-large demonstrates superior performance across all classification settings, particularly with the Structured Image Description technique, achieving 87.95\%, 91.45\%, and 91.88\% accuracy on 6-way, 3-way, and 2-way classification tasks respectively. This represents substantial improvements over its text-only baseline (78.80\%, 86.99\%, 87.23\%).

The performance scaling across model sizes is particularly noteworthy. TinyBERT, despite its compact architecture, shows remarkable improvements with visual information integration, achieving up to 87.36\% accuracy in binary classification compared to its 83.24\% text-only baseline. DistilBERT demonstrates even stronger gains, with its Structured Image Description performance reaching 88.75\% in binary classification, a 3.76 percentage point improvement over its text-only counterpart. BERT-base follows a similar pattern, achieving 89.21\% with Structured Image Description versus 85.09\% with text alone.

Across all models, certain patterns emerge in prompting strategy effectiveness:
\begin{itemize}\setlength{\parskip}{0pt}\setlength{\itemsep}{0pt}
\item Structured Image Description consistently outperforms other techniques, suggesting that well-organized, natural language descriptions are most effective for transformer models
\item Simple Image Description and List of Objects show strong performance despite their simplicity, often achieving within 1-2 percentage points of the best results
\item Scene Graph Analysis performs particularly well with larger models (87.91\% with RoBERTa on 6-way classification) but shows diminished returns with smaller architectures
\item Relational Mapping and Inconsistency Detection, while sophisticated, generally underperform simpler techniques, possibly due to the challenge of effectively encoding complex spatial and visual relationships in text
\end{itemize}

The performance gap between different prompting strategies widens with task complexity, becoming most pronounced in 6-way classification. For instance, RoBERTa-large shows a 20.51 percentage point spread between its best (87.95\%) and worst (67.44\%) prompting strategies in 6-way classification, compared to only 4.72 points in binary classification. This suggests that sophisticated prompting becomes increasingly crucial as the classification task grows more complex.

\begin{table*}
    \centering
    \begin{tabular}{p{4.5cm}llccc}
        \toprule
        \textbf{Model} & \textbf{Prompting Technique} & \textbf{6-way} & \textbf{3-way} & \textbf{2-way} \\
        \midrule
        \multirow{7}{*}{\makecell[l]{\texttt{huawei-noah/}\\
\texttt{TinyBERT\_General\_4L\_312D}}}
            & Text Only & 70.29\% & 82.84\% & 83.24\% \\
            \cmidrule{2-5}
            & Text + List of Objects & 77.02\% & 82.84\% & 85.05\% \\
            & Text + Simple Image Description & 77.71\% & 85.28\% & 85.17\% \\
            & Text + Structured Image Description & 80.08\% & 86.78\% & 87.36\% \\
            & Text + Relational Mapping & 73.96\% & 81.23\% & 84.14\% \\
            & Text + Inconsistency Detection & 75.74\% & 81.06\% & 81.89\% \\
            & Text + Scene Graph Analysis & 74.91\% & 81.94\% & 85.05\% \\
        \midrule
        \multirow{8}{*}{\texttt{distilbert-base-uncased}}
            & Text Only & 75.01\% & 84.83\% & 84.99\% \\
            \cmidrule{2-5}
            & Text + List of Objects & 82.09\% & 88.92\% & 88.50\% \\
            & Text + Simple Image Description & 83.09\% & 87.24\% & 88.69\% \\
            & Text + Structured Image Description & 83.98\% & 88.38\% & 88.75\% \\
            & Text + Relational Mapping & 80.94\% & 86.28\% & 86.91\% \\
            & Text + Inconsistency Detection & 80.55\% & 85.77\% & 85.66\% \\
            & Text + Scene Graph Analysis & 83.86\% & 87.50\% & 87.64\% \\
        \midrule
        \multirow{8}{*}{\texttt{bert-base-uncased}}
            & Text Only & 79.50\% & 84.09\% & 85.09\% \\
            \cmidrule{2-5} 
            & Text + List of Objects & 83.10\% & 87.12\% & 87.45\% \\
            & Text + Simple Image Description & 83.34\% & 88.16\% & 88.57\% \\
            & Text + Structured Image Description & 85.25\% & 88.93\% & 89.21\% \\
            & Text + Relational Mapping & 81.34\% & 86.48\% & 86.43\% \\
            & Text + Inconsistency Detection & 80.95\% & 85.75\% & 85.95\% \\
            & Text + Scene Graph Analysis & 85.24\% & 88.04\% & 88.35\% \\
        \midrule
        \multirow{7}{*}{\texttt{FacebookAI/roberta-large}}
            & Text Only & 78.80\% & 86.99\% & 87.23\% \\
            \cmidrule{2-5}
            & Text + List of Objects & 86.60\% & 89.57\% & 89.92\% \\
            & Text + Simple Image Description & 86.99\% & 90.77\% & 90.74\% \\
            & Text + Structured Image Description & \textbf{87.95\%} & \textbf{91.45\%} & \textbf{91.88\%} \\
            & Text + Relational Mapping & 85.05\% & 89.41\% & 89.49\% \\
            & Text + Inconsistency Detection & 67.44\% & 88.10\% & 89.01\% \\
            & Text + Scene Graph Analysis & 87.91\% & 87.05\% & 87.16\% \\
        \bottomrule        
    \end{tabular}
    \caption{Classification Performance Comparison (Accuracy) of BERT, DistilBERT, and RoBERTa on Uni-Fakeddit-55k}
    \label{tab:classification-results}
\end{table*}

DeBERTa-v3-large demonstrates superior performance across all prompting techniques (Table \ref{tab:deberta-results}), with the Structured Image Description strategy achieving the highest accuracy of 92.52\%. Notable performance gains are also observed with Simple Image Description (92.06\%) and Scene Graph Analysis (92.15\%), suggesting that DeBERTa effectively leverages complex visual information when presented in textual form. The model shows significant improvement over its text-only baseline (87.64\%), with an increase of 4.88 percentage points using Structured Image Description. Even the simplest approach, List of Objects, yields a modest improvement to 88.08\%, while more sophisticated techniques like Relational Mapping achieve 90.72\%. Interestingly, Inconsistency Detection shows minimal gains (87.85\%), possibly due to the complexity of translating visual inconsistencies into text that the model can effectively process.

\begin{table*}
    \centering
    \begin{tabular}{llccc}
        \toprule
        \textbf{Model} & \textbf{Prompting Technique} & \textbf{Accuracy} \\
        \midrule
        \multirow{7}{*}{\texttt{microsoft/deberta-v3-large}}
            & Text Only & 87.64\% \\
            \cmidrule{2-3}
            & Text + List of Objects & 88.08\% \\
            & Text + Simple Image Description & 92.06\% \\
            & Text + Structured Image Description & \textbf{92.52\%} \\
            & Text + Relational Mapping & 90.72\% \\
            & Text + Inconsistency Detection & 87.85\% \\
            & Text + Scene Graph Analysis & 92.15\% \\
        \end{tabular}
    \caption{Binary Classification Accuracy of DeBERTa on Uni-Fakeddit-55k}
    \label{tab:deberta-results}
\end{table*}

\section{Training Costs}
\label{appendix:training-costs}
The training costs for various language models were estimated based on Google Colab pricing of \$10 per 100 Compute Units (CUs). Table \ref{tab:model-training-costs} provides a detailed breakdown of the training times and associated costs.

\begin{table}[ht]
    \centering
    \begin{tabular}{p{2.0 cm}lc}
    \hline
    \textbf{Model} & \textbf{Training Time} & \textbf{Cost} \\
    \hline
    BERT & 35 minutes & \$0.32 \\
    DistilBERT & 22 minutes & \$0.282 \\
    TinyBERT & 11 minutes & \$0.1465 \\
    RoBERTa & 105 minutes & \$1.48 \\
    DeBERTa & 135 minutes & \$1.91 \\
    \hline
    \end{tabular}
    \caption{Language Model Training Costs on Google Colab}
    \label{tab:model-training-costs}
\end{table}

These cost estimates provide insights into the computational resources required for training different language models, highlighting the variations in training complexity and duration.

\section{Selection of Gemini 1.5 Pro for Vision-to-Text Translation}
\label{appendix:gemini-rationale}

Our evaluation of various Vision-Language Models (VLMs) led to the selection of Gemini 1.5 Pro as the primary model for vision-to-text translation. This choice was driven by several key factors:

\subsection{Technical Advantages}
\begin{itemize}\setlength{\parskip}{0pt}\setlength{\itemsep}{0pt}
\item \textbf{Context Window:} With a 2-million token context window, Gemini 1.5 Pro can handle images of varying complexity and generate detailed descriptions without truncation
\item \textbf{Response Quality:} Produces more consistent and detailed outputs compared to open-source alternatives
\item \textbf{Flexibility:} Supports multiple prompting strategies without triggering content restrictions
\end{itemize}

\subsection{Limitations of Alternatives}

\subsubsection{Llama 3.2 Vision}
\begin{itemize}\setlength{\parskip}{0pt}\setlength{\itemsep}{0pt}
\item Built-in guardrails in the mlx-community/Llama-3.2-11B-Vision-Instruct-8bit model frequently block outputs for our prompting tasks
\item Lower performance in zero-shot classification (63.92\% vs. base accuracy)
\item More restrictive in handling complex prompting strategies
\end{itemize}

\subsubsection{GPT-4o}
\begin{itemize}\setlength{\parskip}{0pt}\setlength{\itemsep}{0pt}
\item Limited to 128K token context window
\item Significantly higher operational costs
\item Less suitable for large-scale deployment
\end{itemize}

\subsection{Accessibility Considerations}
\begin{itemize}\setlength{\parskip}{0pt}\setlength{\itemsep}{0pt}
\item \$300 free trial credit enables initial deployment without significant investment
\item Cost-effective for both research and small-scale applications
\item Lower per-token costs compared to GPT-4o
\end{itemize}

The combination of superior technical capabilities, fewer restrictions, and better accessibility makes Gemini 1.5 Pro the optimal choice for our framework's vision-to-text translation component. Its performance characteristics and cost structure align well with our goal of democratizing fake news detection capabilities.

\section{Prompting Techniques and Examples of Uni-Fakeddit-55k Dataset }
\label{appendix:prompts}

This appendix provides detailed information about our six prompting techniques, including the exact prompts used and representative examples for each strategy.

\subsection{List of Objects Strategy}
\label{appendix:list-objects}
\subsubsection{Prompt Template}
\begin{quote}
Analyze the image and list all clearly visible objects and elements. Return a comma-separated list of distinct, identifiable objects. Focus on physical objects, not interpretations or actions. Be specific but concise in naming objects.
\end{quote}
\subsubsection{Example Outputs}
\begin{figure}[ht]

\centering
    \begin{subfigure}[b]{1.0\columnwidth}
        \includegraphics[width=\columnwidth]{Figures/example1\_objects}
        \caption{Input Image 1: Police incident scene}
        \label{fig:list-of-objects-1}
    \end{subfigure}
\textbf{Caption:} Use of deadly force by police disappears on Richmond streets

\textbf{Model Output:} Car exhaust pipe, Car bumper, Car body, Gun shadow, Hand shadow, Reflection of a person, Reflection of a vehicle, Reflection of trees/objects, Road reflection

\vspace{0.5cm}

\begin{subfigure}[b]{1.0\columnwidth}
    \includegraphics[width=\columnwidth]{Figures/example2\_objects}
    \caption{Input Image 2: Theme park concept art}
    \label{fig:list-of-objects-2}
\end{subfigure}

\textbf{Caption:} Viacom's proposed Nickelodeon-themed floating resort to be situated at Coron Palawan

\textbf{Model Output:} Roller coaster, roller coaster track, glove-shaped entrance, sign reading "Glove World", glove-shaped Ferris wheel, glove-shaped ride, fence, glove-shaped topiary, light post with flag, SpongeBob SquarePants, Patrick Star, road/sidewalk, jellyfish, glove decorations, red and white striped tent/awning
\end{figure}

This strategy focuses on creating an objective inventory of visible objects without interpretation or context. The comma-separated format ensures consistent parsing and processing in downstream tasks. Figures \ref{fig:list-of-objects-1} and \ref{fig:list-of-objects-2} present two sample inputs and the respective model outputs for this prompting strategy.

\subsection{Simple Image Description Strategy}
\label{appendix:simple-desc}
\subsubsection{Prompt Template}
\begin{quote}
Describe this image in exactly two sentences. Return only the description, with no additional text or explanation.
\end{quote}
\subsubsection{Example Outputs}
\begin{figure}[ht]
\centering
\begin{subfigure}[b]{1.0\columnwidth}
    \includegraphics[width=\columnwidth]{Figures/example2\_simple}
    \caption{Input Image 1: UN press conference}
    \label{fig:simple-image-desc-1}
\end{subfigure}
\textbf{Caption:} U.S., Britain, France, others skip nuclear weapons ban treaty talks

\textbf{Model Output:} Nikki Haley, then-U.S. Ambassador to the United Nations, speaks at a podium, surrounded by other officials.  They appear to be at a press conference or similar official event.

\vspace{0.5cm}

\begin{subfigure}[b]{1.0\columnwidth}
    \includegraphics[width=\columnwidth]{Figures/example1\_simple}
    \caption{Input Image 2: Teapot modification}
    \label{fig:simple-image-desc-2}
\end{subfigure}

\textbf{Caption:} Use plastic tubing to make a more accurate teapot spout

\textbf{Model Output:} A small, white teapot sits on a brown table. The teapot's lid is slightly askew, revealing a stained interior, and the spout has a piece of clear tape affixed to it.
\end{figure}

This strategy focuses on generating concise, factual descriptions of image content. The two-sentence constraint ensures consistent length while allowing for both primary subject description and contextual details. Two sample inputs and their corresponding model outputs for this prompting strategy are illustrated in Figures \ref{fig:simple-image-desc-1} and \ref{fig:simple-image-desc-2}.

\subsection{Structured Image Description Strategy}

\label{appendix:structured-desc}
\subsubsection{Prompt Template}
\begin{quote}
Provide exactly two sentences about this image:
First sentence: State only the observable facts - who and what is in the image, and where it takes place. Focus solely on what can be directly seen.
Second sentence: Interpret the context - explain the likely purpose, situation, or story behind what's shown, including any relevant how or why elements.
Return only these two sentences, with no additional text.
\end{quote}
\subsubsection{Example Outputs}
\begin{figure}[ht]
\centering
\begin{subfigure}[b]{1.0\columnwidth}
    \includegraphics[width=\columnwidth]{Figures/example1\_structured.jpg}
    \caption{Input Image 1: Walt Disney at Disneyland}
    \label{fig:structured-image-desc-1}
\end{subfigure}
\textbf{Caption:} Walt Disney, deep in thought, after everyone's left the park

\textbf{Model Output:} Walt Disney sits on a green bench, facing a large Victorian-style building with a clock tower, situated in a paved plaza with trees. This photo likely captures Disney observing his newly opened Disneyland, perhaps reflecting on its creation or planning future additions and improvements.

\vspace{0.5cm}

\begin{subfigure}[b]{1.0\columnwidth}
    \includegraphics[width=\columnwidth]{Figures/example2\_structured.jpg}
    \caption{Input Image 2: Octopus underwater}
    \label{fig:structured-image-desc-2}
\end{subfigure}

\textbf{Caption:} This octopus discovered a coconut on the bottom of the ocean you won't believe what happened next

\textbf{Model Output:} An octopus, partially camouflaged against a rocky, green seabed, holds a large crab in its tentacles underwater. The octopus is likely hunting and has captured the crab as prey, using its tentacles to restrain it before consuming it.
\end{figure}

This strategy employs a two-part structure to separate objective observations from contextual interpretation. The first sentence focuses on directly observable elements, while the second sentence provides deeper context and interpretation, enabling a more comprehensive understanding of the image content. The model outputs for two example inputs using this prompting strategy are depicted in Figures \ref{fig:structured-image-desc-1} and \ref{fig:structured-image-desc-2}.

\subsection{Relational Mapping Strategy}
\label{appendix:relational}

\subsubsection{Prompt Template}
\begin{quote}
Analyze the spatial and interactive relationships between objects in this image. Return a JSON response with the following structure:

\begin{verbatim}
{
"objects": [
{
"id": "unique_number",
"name": "object_name",
"location": "general_location_in_image"
}
],
"relationships": [
{
"subject_id": "id_of_first_object",
"relation": "type_of_relationship",
"object_id": "id_of_second_object",
"confidence": "float_between_0_and_1"
}
]
}
\end{verbatim}

Include ONLY clearly visible relationships. For each relationship, assign a confidence score.
\end{quote}

\subsubsection{Example Outputs}

\begin{figure}[ht]
    \centering
    \begin{subfigure}[b]{1.0\columnwidth}
        \includegraphics[width=\columnwidth]{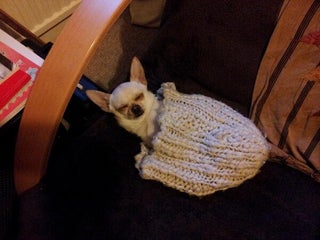}
        \caption{Input Image 1: Pet portrait}
        \label{fig:relational-map-1}
    \end{subfigure}
    
    \textbf{Caption:} Chihuahua in a hat
    
    \textbf{Model Output:}
    \begin{lstlisting}[basicstyle=\small\ttfamily]
{"objects": [{"id": "1","name": "chihuahua","location": "center"},{"id": "2","name": "sweater","location": "center"},{"id": "3","name": "chair","location": "everywhere"},{"id": "4","name": "cushion","location": "underneath dog"},{"id": "5","name": "blanket/pillow","location": "right side"},{"id": "6","name": "shelf/table edge","location": "left side"},{"id": "7","name": "red and white object","location": "top left corner"}],"relationships": [{"subject_id": "1","relation": "wearing","object_id": "2","confidence": 1.0},{"subject_id": "1","relation": "lying on","object_id": "3","confidence": 1.0},{"subject_id": "1","relation": "lying on","object_id": "4","confidence": 1.0},{"subject_id": "3","relation": "next to","object_id": "5","confidence": 1.0},{"subject_id": "3","relation": "partially occluding","object_id": "6","confidence": 0.9},{"subject_id": "6","relation": "supporting","object_id": "7","confidence": 0.8}]}
    \end{lstlisting}
\end{figure}
\begin{figure}
    \begin{subfigure}[b]{1.0\columnwidth}
        \includegraphics[width=\columnwidth]{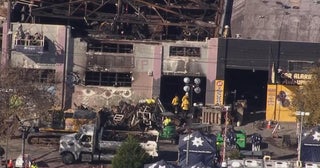}
        \caption{Input Image 2: Fire aftermath}
        \label{fig:relational-map-2}
    \end{subfigure}
    
    \textbf{Caption:} Inspectors tried to access Oakland warehouse before deadly fire
    
    \textbf{Model Output:}
    \begin{lstlisting}[basicstyle=\small\ttfamily]
{"objects": [{"id": 1,"name": "burned building","location": "center"},{"id": 2,"name": "debris pile","location": "bottom center"},{"id": 3,"name": "truck","location": "bottom left"},{"id": 4,"name": "firefighters","location": "center"},{"id": 5,"name": "construction equipment","location": "bottom right"},{"id": 6,"name": "white tent","location": "bottom center"},{"id": 7,"name": "adjacent building","location": "right"}],"relationships": [{"subject_id": 2,"relation": "in front of","object_id": 1,"confidence": 1.0},{"subject_id": 3,"relation": "next to","object_id": 2,"confidence": 0.9},{"subject_id": 4,"relation": "inspecting","object_id": 1,"confidence": 0.8},{"subject_id": 4,"relation": "near","object_id": 2,"confidence": 0.7},{"subject_id": 5,"relation": "parked near","object_id": 1,"confidence": 0.9},{"subject_id": 6,"relation": "in front of","object_id": 1,"confidence": 0.8},{"subject_id": 7,"relation": "attached to","object_id": 1,"confidence": 0.9}]}
    \end{lstlisting}
\end{figure}

This strategy generates structured representations of spatial and interactive relationships between objects in the image. The JSON format enables precise encoding of object locations and their relationships, with confidence scores indicating the certainty of each relationship identification. Figures \ref{fig:relational-map-1} and \ref{fig:relational-map-2} showcase the model's responses to two representative inputs when employing this prompting approach.

\subsection{Inconsistency Detection Strategy}
\label{appendix:inconsistency}

\subsubsection{Prompt Template}
\begin{quote}
Analyze this image for potential manipulation indicators or inconsistencies. Return a JSON response structured as follows:

\begin{lstlisting}
{
"lighting_analysis": {
"inconsistencies": [
{
"description": "Detailed description of the lighting inconsistency",
"location": "Where in the image this occurs",
"confidence": float between 0-1,
"affected_objects": ["list", "of", "affected", "objects"]
}],
"overall_lighting_coherence": float between 0-1
},
"perspective_analysis": {
"inconsistencies": [
{
"description": "Description of perspective or geometric anomaly",
"location": "Where in the image this occurs",
"confidence": float between 0-1,
"affected_objects": ["list", "of", "affected", "objects"]
}],
"overall_perspective_coherence": float between 0-1
},
"boundary_analysis": {
"suspicious_edges": [
{
"object": "Name of object",
"description": "Description of boundary anomaly",
"location": "Where in the image this occurs",
"confidence": float between 0-1
}],
"overall_edge_quality": float between 0-1
},
"resolution_analysis": {
"inconsistencies": [{
"object": "Name of object",
"description": "Description of resolution mismatch",
"relative_to": "What it's inconsistent with",
"confidence": float between 0-1
}],
"overall_resolution_coherence": float between 0-1
},
"metadata_analysis": {
"jpeg_artifacts": boolean,
"compression_inconsistencies": boolean,
"noise_patterns": ["list", "of", "suspicious", "patterns"]
},
"summary": {
"manipulation_likelihood": float between 0-1,
"most_suspicious_elements": ["list", "of", "concerning", "elements"],
"overall_assessment": "Brief summary of findings"}}
\end{lstlisting}

Focus on identifying concrete, observable inconsistencies rather than speculation. Provide specific locations and descriptions for each identified anomaly.
\end{quote}

\subsubsection{Example Outputs}

\begin{figure}[ht]
    \centering
    \begin{subfigure}[b]{0.9\columnwidth}
        \includegraphics[width=0.9\columnwidth]{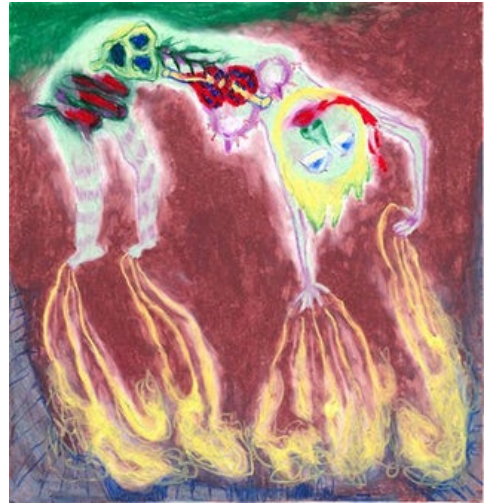}
        \caption{Input Image 1: Artistic rendering}
        \label{fig:inconsistency-1}
    \end{subfigure}
    
    \textbf{Caption:} Flooded boiler room in Washington a few filters over a dog
    
    \textbf{Model Output:}
    \begin{lstlisting}[basicstyle=\small\ttfamily]
{"lighting_analysis": {"inconsistencies": [],"overall_lighting_coherence": 0.9},"perspective_analysis": {"inconsistencies": [{"description": "The figure's limbs and body seem to defy normal anatomical proportions and perspective, appearing stretched and distorted.","location": "Throughout the figure","confidence": 0.8,"affected_objects": ["Figure"]}],"overall_perspective_coherence": 0.5},"boundary_analysis": {"suspicious_edges": [],"overall_edge_quality": 0.9},"resolution_analysis": {"inconsistencies": [],"overall_resolution_coherence": 0.9},"metadata_analysis": {"jpeg_artifacts": false,"compression_inconsistencies": false,"noise_patterns": []},"summary": {"manipulation_likelihood": 0.1,"most_suspicious_elements": [],"overall_assessment": "The image is likely an artistic creation with stylized, non-realistic proportions. No clear signs of image manipulation were detected. The distorted anatomy is consistent with artistic expression rather than technical manipulation."}}
    \end{lstlisting}
    
\end{figure}

\begin{figure}
    \begin{subfigure}[b]{1.0\columnwidth}
        \includegraphics[width=\columnwidth]{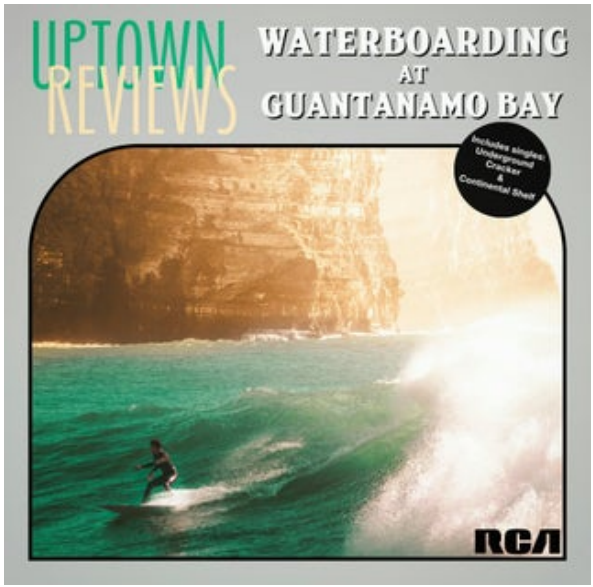}
        \caption{Input Image 2: Restaurant review}
        \label{fig:inconsistency-2}
    \end{subfigure}

    \textbf{Caption:} Uptown Reviews - Waterboarding at Guantanamo Bay
    
    \textbf{Model Output:}
    \begin{lstlisting}[basicstyle=\small\ttfamily]
{"lighting_analysis": {"inconsistencies": [],"overall_lighting_coherence": 0.95},"perspective_analysis": {"inconsistencies": [],"overall_perspective_coherence": 0.95},"boundary_analysis": {"suspicious_edges": [],"overall_edge_quality": 0.95},"resolution_analysis": {"inconsistencies": [],"overall_resolution_coherence": 0.95},"metadata_analysis": {"jpeg_artifacts": false,"compression_inconsistencies": false,"noise_patterns": []},"summary": {"manipulation_likelihood": 0.05,"most_suspicious_elements": [],"overall_assessment": "No obvious signs of image manipulation were detected. The lighting, perspective, boundaries, and resolution appear consistent throughout the image. There are no discernible compression artifacts or unusual noise patterns."}}
    \end{lstlisting}
\end{figure}

This strategy performs a comprehensive analysis of potential image manipulations by examining multiple technical aspects including lighting, perspective, boundaries, resolution, and metadata. The structured JSON output provides detailed assessments of each aspect with confidence scores and specific locations of any detected anomalies. Figures \ref{fig:inconsistency-1} and \ref{fig:inconsistency-2} show two examples of inputs and outputs using this prompting strategy.

\subsection{Scene Graph Analysis Strategy}
\label{appendix:scene-graph}

\subsubsection{Prompt Template}
\begin{quote}
Analyze this image to create a detailed scene graph with metadata. Return a JSON response structured as follows:

\begin{lstlisting}
{
"primary_subject": {
"description": "Detailed description of the main subject/focus",
"confidence": float between 0-1,
"typical_context": boolean,
"context_notes": "Explanation of context typicality"
},
"scene_elements": [
{
"object": "Name of the object or element",
"location": "Location in the image",
"confidence": float between 0-1,
"relationships": [
{
"related_to": "Name of the related object",
"relationship_type": "Type of relationship",
"confidence": float between 0-1,
"description": "Brief description of relationship"
}
],
"inconsistencies": [
{
"type": "Type of inconsistency",
"description": "Description of visual inconsistency",
"severity": float between 0-1
}]}],
"metadata_analysis": {
"image_quality": float between 0-1,
"quality_factors": {
"resolution": float between 0-1,
"clarity": float between 0-1,
"lighting": float between 0-1
},
"potential_manipulations": [...],
"technical_artifacts": [...]
},
"analysis_summary": {
"scene_complexity": float between 0-1,
"manipulation_likelihood": float between 0-1,
"overall_consistency": float between 0-1,
"key_observations": ["list", "of", "findings"]
}}
\end{lstlisting}

Focus on creating accurate relationships between elements and identifying any unusual or inconsistent aspects of the scene.
\end{quote}

\subsubsection{Example Outputs}

\begin{figure}[ht]
    %\centering
    \begin{subfigure}[b]{0.9\columnwidth}
        \includegraphics[width=0.9\columnwidth]{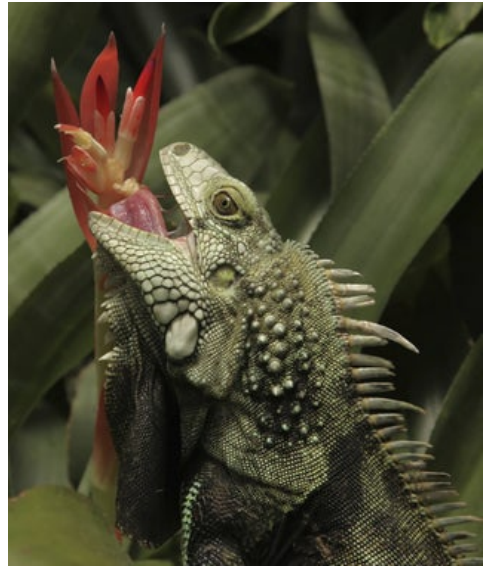}
        \caption{Input Image 1: Wildlife photograph}
        \label{fig:scene-graph-1}
    \end{subfigure}
    
    \textbf{Caption:} Iguana eating a flower
    
    \textbf{Model Output:}
    \begin{lstlisting}[basicstyle=\small\ttfamily]
{"primary_subject":{"description":"A green iguana is consuming a red, spiky flower.","confidence":0.95,"typical_context": true,"context_notes": "Iguanas are known to eat flowers and vegetation,so this behavior is within their natural diet."},"scene_elements":[{"object": "Iguana","location": "Center","confidence": 0.99,"relationships":[{"related_to": "Flower","relationship_type": "Eating","confidence": 0.95,"description":"The iguana's mouth is open and around the flower, indicating it is consuming it."},{"related_to": "Leaves","relationship_type": "Near","confidence": 0.90,"description": "The iguana is surrounded by green leaves, suggesting a natural habitat or enclosure."}],"inconsistencies":[]},{"object": "Flower","location": "Left-center","confidence":0.90,"relationships":[{"related_to": "Iguana","relationship_type": "Being eaten by","confidence": 0.95,"description":"The flower is in the iguana's mouth."}],"inconsistencies":[]},{"object":"Leaves","location":"Background","confidence": 0.95,"relationships": [{"related_to": "Iguana","relationship_type": "Surrounding","confidence":0.90,"description": "The leaves form the background and immediate environment of the iguana."}] ,"inconsistencies": []}],"metadata_analysis":{"image_quality": 0.85,"quality_factors": {"resolution": 0.80,"clarity": 0.90,"lighting": 0.80},"potential_manipulations":[],"technical_artifacts": [{"type":"Slight noise","location": "Darker areas of the leaves","severity": 0.1}]},"analysis_summary": {"scene_complexity":0.5,"manipulation_likelihood":0.05,"overall_consistency": 0.95,
\end{lstlisting}
\end{figure}

\begin{figure}
\begin{lstlisting}[basicstyle=\small\ttfamily]
"key_observations":["The image depicts a natural interaction between an iguana and a flower in a leafy environment.","The image quality is good with slight noise in darker areas.","No obvious signs of manipulation were detected."]}}
\end{lstlisting}
\end{figure}

\begin{figure}
    \begin{subfigure}[b]{0.9\columnwidth}
        \includegraphics[width=0.9\columnwidth]{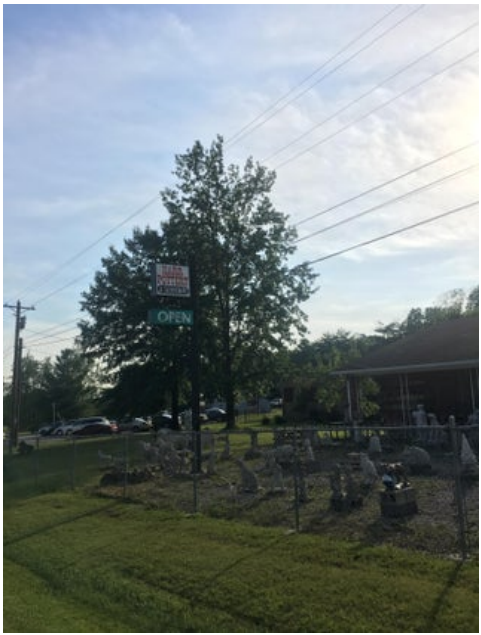}
        \caption{Input Image 2: Yard decoration}
        \label{fig:scene-graph-2}
    \end{subfigure}
    
    \textbf{Caption:} The name of this shop is almost Harry Potter
    
    \textbf{Model Output:}
    \begin{lstlisting}[basicstyle=\small\ttfamily]
{"primary_subject": {"description": "A yard decorated with concrete statues and figures.","confidence": 0.9,"typical_context": false,"context_notes": "While yards can have decorations, the sheer number and apparent focus on concrete statues makes this context less typical than a yard with a few gnomes or other common decorations."},"scene_elements": [{"object": "Concrete statues","location": "Center, foreground","confidence": 0.95,"relationships": [{"related_to": "Yard","relationship_type": "Located in","confidence": 1.0,"description": "The statues are placed throughout the yard."},{"related_to": "Fence","relationship_type": "Enclosed by","confidence": 0.8,"description": "A chain-link fence appears to enclose the area where the statues are displayed."}],"inconsistencies": []},{"object": "Yard","location": "Foreground","confidence": 1.0, 
\end{lstlisting}
\end{figure}

\begin{figure}[!htbp]
\begin{lstlisting}[basicstyle=\small\ttfamily]
"inconsistencies": []},{"object": "Cars","location": "Left, middleground","confidence": 0.9,"relationships": [],"inconsistencies": []},{"object": "Trees","location": "Background and throughout the scene","confidence": 0.95,"relationships": [],"inconsistencies": []},{"object": "Sign","location": "Center-left, above the statues","confidence": 0.8,"relationships": [{"related_to": "Concrete statues","relationship_type": "Relates to","confidence": 0.7,"description": "Likely a sign advertising the sale or creation of the statues."}],"inconsistencies": []},{"object": "Fence","location": "Surrounding the statues","confidence": 0.8,"relationships": [{"related_to": "Concrete Statues","relationship_type": "Encloses","confidence": 0.9,"description": "The fence likely serves to enclose the statue display area."}],"inconsistencies": []},{"object": "Power lines","location": "Across the image","confidence": 0.9,"relationships": [],"inconsistencies": []}],"metadata_analysis": {"image_quality": 0.75,"quality_factors": {"resolution": 0.8,"clarity": 0.8,"lighting": 0.7},"potential_manipulations": [],"technical_artifacts": []},"analysis_summary": {"scene_complexity": 0.6,"manipulation_likelihood": 0.05,"overall_consistency": 0.95,"key_observations": ["The primary subject is a collection of concrete statues in a yard.  The scene is generally consistent and appears to be an unedited photo."]}}
\end{lstlisting}

\end{figure}

This strategy creates comprehensive scene graphs that capture both the physical layout and the semantic relationships between objects in the image. The structured output includes confidence scores for detections, relationship descriptions, and detailed metadata analysis, enabling rich understanding of scene composition and potential inconsistencies. To demonstrate the effectiveness of this prompting strategy, Figures \ref{fig:scene-graph-1} and \ref{fig:scene-graph-2} display two sample inputs alongside their respective model-generated outputs.

\section{Training Hyperparameters}
\label{appendix:training-parameters}

This appendix provides detailed information about the training configuration and hyperparameters used for each model in our experiments.
\subsection{BERT Configuration}
\label{appendix:bert-config}
\subsubsection{Model Architecture}
\begin{itemize}\setlength{\parskip}{0pt}\setlength{\itemsep}{0pt}
\item Base Model: \texttt{bert-base-uncased}
\item Number of Labels: 6 (multi-class classification)
\item Maximum Sequence Length: 128 tokens
\item Model Size: 110M parameters
\end{itemize}
\subsubsection{Training Parameters}
\begin{itemize}\setlength{\parskip}{0pt}\setlength{\itemsep}{0pt}
\item Batch Size: 8 (per device)
\item Number of Epochs: 5
\item Optimizer: AdamW
\item Learning Rate: Default (5e-5)
\item Weight Decay: 0.01
\item Training/Validation Split: 70\%/30\%
\end{itemize}
\subsubsection{Data Processing}
\begin{itemize}\setlength{\parskip}{0pt}\setlength{\itemsep}{0pt}
\item Text Input Format: \texttt{[TEXT] + [OBJECTS]} concatenation
\item Tokenization: BERT tokenizer with truncation
\item Padding Strategy: Max length (128 tokens)
\item Random Seed: 42 (for reproducibility)
\end{itemize}
\subsubsection{Training Strategy}
\begin{itemize}\setlength{\parskip}{0pt}\setlength{\itemsep}{0pt}
\item Evaluation Strategy: Per epoch
\item Save Strategy: Per epoch
\item Best Model Selection: Based on macro F1-score
\item Early Stopping: Disabled
\item Gradient Accumulation Steps: 1
\item Mixed Precision Training: Enabled (FP16)
\end{itemize}
\subsubsection{Evaluation Metrics}
\begin{itemize}\setlength{\parskip}{0pt}\setlength{\itemsep}{0pt}
\item Primary Metrics: Accuracy, Macro F1-score
\item Secondary Metrics: Precision (macro and weighted)
\item Per-class Metrics: Precision and F1-score
\item Zero-division Handling: Set to 0 for undefined cases
\end{itemize}
\subsubsection{Hardware Configuration}
\begin{itemize}\setlength{\parskip}{0pt}\setlength{\itemsep}{0pt}
\item Computing Platform: RTX 4060 Laptop GPU
\item CUDA Version: 12.6
\item Memory Utilization: 0.4GB VRAM (peak)
\item Training Time: Approximately 1.16 hours
\end{itemize}
The training configuration was optimized for the specific characteristics of our dataset and hardware constraints. All experiments were conducted using the Hugging Face Transformers library (version 4.48.3) and PyTorch (version 2.6.0+cu126).

\subsection{TinyBERT Configuration}
\label{appendix:tinybert-config}

\subsubsection{Model Architecture}
\begin{itemize}\setlength{\parskip}{0pt}\setlength{\itemsep}{0pt}
\item Base Model: \texttt{huawei-noah/TinyBERT\_General\_4L\_312D}
\item Number of Labels: 6 (multi-class classification)
\item Maximum Sequence Length: 512 tokens
\item Model Size: 14.5M parameters
\item Architecture: 4-layer, 312-dimensional compressed BERT
\end{itemize}

\subsubsection{Training Parameters}
\begin{itemize}\setlength{\parskip}{0pt}\setlength{\itemsep}{0pt}
\item Batch Size: 16 (per device)
\item Number of Epochs: 5
\item Optimizer: AdamW
\item Learning Rate: 2e-5
\item Weight Decay: 0.01
\item Warmup Ratio: 0.1
\item Training/Validation Split: 70\%/30\%
\end{itemize}

\subsubsection{Data Processing}
\begin{itemize}\setlength{\parskip}{0pt}\setlength{\itemsep}{0pt}
\item Text Input Format: \texttt{[TEXT] + [OBJECTS]} concatenation
\item Tokenization: TinyBERT tokenizer with truncation
\item Padding Strategy: Max length (512 tokens)
\item Random Seed: 42 (for reproducibility)
\end{itemize}

\subsubsection{Training Strategy}
\begin{itemize}\setlength{\parskip}{0pt}\setlength{\itemsep}{0pt}
\item Evaluation Strategy: Every 1000 steps
\item Save Strategy: Every 1000 steps
\item Best Model Selection: Based on macro F1-score
\item Early Stopping: Disabled
\item Gradient Accumulation Steps: 2
\item Mixed Precision Training: Enabled (FP16)
\item Number of Data Loading Workers: 4
\end{itemize}

\subsubsection{Evaluation Metrics}
\begin{itemize}\setlength{\parskip}{0pt}\setlength{\itemsep}{0pt}
\item Primary Metrics: Accuracy, Macro F1-score
\item Secondary Metrics: Precision (macro and weighted)
\item Per-class Metrics: Precision and F1-score
\item Zero-division Handling: Set to 0 for undefined cases
\item Evaluation Frequency: Every 1000 training steps
\end{itemize}

\subsubsection{Hardware Configuration}
\begin{itemize}\setlength{\parskip}{0pt}\setlength{\itemsep}{0pt}
\item Computing Platform: RTX 4060 Laptop GPU
\item CUDA Version: 12.6
\item Memory Utilization: 0.2GB VRAM (peak)
\item Training Time: Approximately 0.66 hours
\end{itemize}

Key differences from BERT include the use of a compressed 4-layer architecture, larger batch size (16 vs 8), more frequent evaluation (every 1000 steps vs per epoch), and higher learning rate (2e-5 vs 5e-5). These modifications leverage TinyBERT's efficient architecture while maintaining competitive performance.

\subsection{DistilBERT Configuration}
\label{appendix:distilbert-config}

\subsubsection{Model Architecture}
\begin{itemize}\setlength{\parskip}{0pt}\setlength{\itemsep}{0pt}
\item Base Model: \texttt{distilbert-base-uncased}
\item Number of Labels: 6 (multi-class classification)
\item Maximum Sequence Length: 128 tokens
\item Model Size: 66M parameters
\item Architecture: Knowledge-distilled BERT with 6 layers
\end{itemize}

\subsubsection{Training Parameters}
\begin{itemize}\setlength{\parskip}{0pt}\setlength{\itemsep}{0pt}
\item Batch Size: 8 (per device)
\item Number of Epochs: 5
\item Optimizer: AdamW
\item Learning Rate: Default (5e-5)
\item Weight Decay: 0.01
\item Training/Validation Split: 70\%/30\%
\end{itemize}

\subsubsection{Data Processing}
\begin{itemize}\setlength{\parskip}{0pt}\setlength{\itemsep}{0pt}
\item Text Input Format: \texttt{[TEXT] + [OBJECTS]} concatenation
\item Tokenization: DistilBERT tokenizer with truncation
\item Padding Strategy: Max length (128 tokens)
\item Random Seed: 42 (for reproducibility)
\end{itemize}

\subsubsection{Training Strategy}
\begin{itemize}\setlength{\parskip}{0pt}\setlength{\itemsep}{0pt}
\item Evaluation Strategy: Every 2000 steps
\item Save Strategy: Every 2000 steps
\item Best Model Selection: Based on macro F1-score
\item Early Stopping: Disabled
\item Gradient Accumulation Steps: 1
\item Mixed Precision Training: Not enabled
\end{itemize}

\subsubsection{Evaluation Metrics}
\begin{itemize}\setlength{\parskip}{0pt}\setlength{\itemsep}{0pt}
\item Primary Metrics: Accuracy, Macro F1-score
\item Secondary Metrics: Precision (macro and weighted)
\item Per-class Metrics: Precision and F1-score
\item Zero-division Handling: Set to 0 for undefined cases
\item Evaluation Frequency: Every 2000 training steps
\end{itemize}

\subsubsection{Hardware Configuration}
\begin{itemize}\setlength{\parskip}{0pt}\setlength{\itemsep}{0pt}
\item Computing Platform: RTX 4060 Laptop GPU
\item CUDA Version: 12.6
\item Memory Utilization: 0.4GB VRAM (peak)
\item Training Time: Approximately 1.5 hours
\end{itemize}

DistilBERT represents a middle ground between BERT and TinyBERT, using knowledge distillation to achieve a 40\% size reduction from BERT while maintaining relatively strong performance. The configuration balances efficiency with model capacity, using similar batch sizes to BERT but with more frequent evaluation steps.

\subsection{RoBERTa Configuration}
\label{appendix:roberta-config}

\subsubsection{Model Architecture}
\begin{itemize}\setlength{\parskip}{0pt}\setlength{\itemsep}{0pt}
\item Base Model: \texttt{FacebookAI/roberta-large}
\item Number of Labels: 6 (multi-class classification)
\item Maximum Sequence Length: 512 tokens
\item Model Size: 355M parameters
\item Architecture: 24-layer optimized BERT variant
\end{itemize}

\subsubsection{Training Parameters}
\begin{itemize}\setlength{\parskip}{0pt}\setlength{\itemsep}{0pt}
\item Batch Size: 4 (per device)
\item Effective Batch Size: 16 (with gradient accumulation)
\item Number of Epochs: 5
\item Optimizer: AdamW
\item Learning Rate: 1e-5
\item Weight Decay: 0.01
\item Warmup Ratio: 0.1
\item Training/Validation Split: 70\%/30\%
\end{itemize}

\subsubsection{Data Processing}
\begin{itemize}\setlength{\parskip}{0pt}\setlength{\itemsep}{0pt}
\item Text Input Format: \texttt{[TEXT] + [OBJECTS]} concatenation
\item Tokenization: RoBERTa tokenizer with truncation
\item Padding Strategy: Max length (512 tokens)
\item Random Seed: 42 (for reproducibility)
\end{itemize}

\subsubsection{Training Strategy}
\begin{itemize}\setlength{\parskip}{0pt}\setlength{\itemsep}{0pt}
\item Evaluation Strategy: Every 2000 steps
\item Save Strategy: Every 2000 steps
\item Best Model Selection: Based on macro F1-score
\item Early Stopping: Disabled
\item Gradient Accumulation Steps: 4
\item Mixed Precision Training: Enabled (FP16)
\item Gradient Checkpointing: Enabled
\item Number of Data Loading Workers: 4
\end{itemize}

\subsubsection{Evaluation Metrics}
\begin{itemize}\setlength{\parskip}{0pt}\setlength{\itemsep}{0pt}
\item Primary Metrics: Accuracy, Macro F1-score
\item Secondary Metrics: Precision (macro and weighted)
\item Per-class Metrics: Precision and F1-score
\item Zero-division Handling: Set to 0 for undefined cases
\item Evaluation Frequency: Every 2000 training steps
\end{itemize}

\subsubsection{Hardware Configuration}
\begin{itemize}\setlength{\parskip}{0pt}\setlength{\itemsep}{0pt}
\item Computing Platform: Nvidia A100
\item CUDA Version: 12.4
\item Memory Utilization: 7.2GB VRAM (peak)
\item Training Time: Approximately 1.75 hours
\end{itemize}

RoBERTa-large is one of the two large models we used for the tests, requiring specific optimizations for training on limited hardware. These include reduced batch size, gradient accumulation, gradient checkpointing, and mixed precision training. Despite the computational demands, these optimizations enable effective training while maintaining the model's superior performance characteristics.

\subsection{DeBERTa Configuration}
\label{appendix:deberta-config}

\subsubsection{Model Architecture}
\begin{itemize}\setlength{\parskip}{0pt}\setlength{\itemsep}{0pt}
\item Base Model: \texttt{microsoft/deberta-v3-large}
\item Number of Labels: 6 (multi-class classification)
\item Maximum Sequence Length: 512 tokens
\item Model Size: 400M parameters
\item Architecture: Enhanced BERT with disentangled attention
\end{itemize}

\subsubsection{Training Parameters}
\begin{itemize}\setlength{\parskip}{0pt}\setlength{\itemsep}{0pt}
\item Batch Size: 4 (per device)
\item Effective Batch Size: 16 (with gradient accumulation)
\item Number of Epochs: 3
\item Optimizer: AdamW
\item Learning Rate: 1e-5
\item Weight Decay: 0.01
\item Warmup Ratio: 0.1
\item Training/Validation Split: 70\%/30\%
\end{itemize}

\subsubsection{Data Processing}
\begin{itemize}\setlength{\parskip}{0pt}\setlength{\itemsep}{0pt}
\item Text Input Format: \texttt{[TEXT] + [OBJECTS]} concatenation
\item Tokenization: DeBERTa tokenizer with truncation
\item Padding Strategy: Max length (512 tokens)
\item Random Seed: 42 (for reproducibility)
\end{itemize}

\subsubsection{Training Strategy}
\begin{itemize}\setlength{\parskip}{0pt}\setlength{\itemsep}{0pt}
\item Evaluation Strategy: Every 2000 steps
\item Save Strategy: Every 2000 steps
\item Best Model Selection: Based on macro F1-score
\item Early Stopping: Disabled
\item Gradient Accumulation Steps: 4
\item Mixed Precision Training: Enabled (FP16)
\item Gradient Checkpointing: Enabled
\item Number of Data Loading Workers: 4
\end{itemize}

\subsubsection{Evaluation Metrics}
\begin{itemize}\setlength{\parskip}{0pt}\setlength{\itemsep}{0pt}
\item Primary Metrics: Accuracy, Macro F1-score
\item Secondary Metrics: Precision (macro and weighted)
\item Per-class Metrics: Precision and F1-score
\item Zero-division Handling: Set to 0 for undefined cases
\item Evaluation Frequency: Every 2000 training steps
\end{itemize}

\subsubsection{Hardware Configuration}
\begin{itemize}\setlength{\parskip}{0pt}\setlength{\itemsep}{0pt}
\item Computing Platform: Nvidia A100
\item CUDA Version: 12.4
\item Memory Utilization: 8.7GB VRAM
\item Training Time: Approximately 2 hours
\end{itemize}

As our most advanced model, DeBERTa-v3-large incorporates disentangled attention mechanisms and enhanced mask decoder training, requiring similar memory optimizations to RoBERTa but with shorter training time due to faster convergence. The architecture's sophisticated attention mechanism enables better handling of complex textual relationships while maintaining computational efficiency through gradient checkpointing and mixed precision training.

Initial runs with 5 epochs resulted in overfitting. Subsequently, we switched to 3 epoch runs, which helped cut training time while deliver higher performance. Evaluation strategy is set to 2,000 steps to further reduce computational cost.

\subsection{Evaluation Methodology}
\label{appendix:evaluation}

Our evaluation framework combines both periodic monitoring during training and comprehensive final assessment. During training, we implement regular evaluation checkpoints to track model progress and ensure stable optimization:

\subsubsection{Training-time Evaluation}
Models are evaluated on the test set at regular intervals during training:
\begin{itemize}\setlength{\parskip}{0pt}\setlength{\itemsep}{0pt}
\item TinyBERT: Every 1000 steps
\item BERT and DistilBERT: Every 2000 steps
\item RoBERTa and DeBERTa: Every 2000 steps with gradient accumulation
\end{itemize}

These intermediate evaluations help monitor:
\begin{itemize}\setlength{\parskip}{0pt}\setlength{\itemsep}{0pt}
\item Convergence patterns
\item Potential overfitting
\item Learning rate effectiveness
\item Model stability
\end{itemize}

\subsubsection{Final Evaluation}
The final evaluation is conducted after completing the full training regime:
\begin{itemize}\setlength{\parskip}{0pt}\setlength{\itemsep}{0pt}
\item BERT, DistilBERT, TinyBERT, and RoBERTa: 5 complete epochs
\item DeBERTa: 3 complete epochs (faster convergence observed)
\end{itemize}

For each model, we compute:
\begin{itemize}\setlength{\parskip}{0pt}\setlength{\itemsep}{0pt}
\item Overall accuracy
\item Macro and weighted precision
\item Macro and weighted F1-scores
\item Per-class precision and F1-scores
\end{itemize}

The evaluation metrics are calculated on the held-out test set (30\% of the dataset) using scikit-learn's implementation of precision, F1-score, and accuracy metrics. All models use identical train-test splits (stratified by class) to ensure fair comparison, with results reported as percentages for clarity.

\section{Vision-Language Model Testing Configuration}
\label{appendix:vlm-setup}

This appendix details the setup and configuration used for zero-shot testing with Vision-Language Models (VLMs).

\subsection{Llama 3.2 Vision Testing Setup}
\label{appendix:llama-vision}

\subsubsection{Model Configuration}
\begin{itemize}\setlength{\parskip}{0pt}\setlength{\itemsep}{0pt}
\item Base Model: \texttt{unsloth/Llama-3.2-11B-Vision-Instruct}
\item Model Size: 11B parameters
\item Quantization: 4-bit quantization with NF4 format
\item Double Quantization: Enabled
\item Dtype: BFloat16
\item Device Mapping: Automatic
\end{itemize}

\subsubsection{Prompt Template}
\begin{lstlisting}
Analyze this image and its title: "{title}"

Is this a fake news post or a real news post? 
Respond with either FAKE or REAL, followed by your explanation.

Title: {title}
\end{lstlisting}

\subsubsection{Processing Configuration}
\begin{itemize}\setlength{\parskip}{0pt}\setlength{\itemsep}{0pt}
\item Maximum Generated Tokens: 100
\item Image Format: RGB (converted if necessary)
\item Batch Size: 1 (sequential processing)
\item Memory Optimization: 4-bit quantization
\end{itemize}

\subsubsection{Evaluation Setup}
\begin{itemize}\setlength{\parskip}{0pt}\setlength{\itemsep}{0pt}
\item Dataset Size: 55,000 samples
\item Output Processing: First word extraction (FAKE/REAL)
\item Label Mapping: FAKE → 0, REAL → 1
\item Default Fallback: FAKE (conservative approach)
\item Progress Monitoring: Accuracy reported every 100 samples
\end{itemize}

\subsubsection{Hardware Requirements}
\begin{itemize}\setlength{\parskip}{0pt}\setlength{\itemsep}{0pt}
\item GPU: NVIDIA RTX 4090
\item VRAM Usage: ~12GB (with quantization)
\item Storage: 50GB for model and dataset
\item Processing Time: Approximately 6 hours for full dataset
\end{itemize}

Results from this configuration achieved 63.92\% accuracy on binary classification, demonstrating the model's zero-shot capabilities in fake news detection without any fine-tuning.

\subsection{Video LLaVA Testing Setup}
\label{appendix:video-llava}

\subsubsection{Model Configuration}
\begin{itemize}\setlength{\parskip}{0pt}\setlength{\itemsep}{0pt}
\item Base Model: \texttt{LanguageBind/Video-LLaVA-7B-hf}
\item Model Size: 7B parameters
\item Model Type: Conditional Generation
\item Quantization: None (full precision)
\item Device Mapping: Single GPU
\end{itemize}

\subsubsection{Prompt Template}
\begin{lstlisting}
USER: <image> Given this image and its title: '{title}', 
is this a fake news post or a real news post? 
Respond with either FAKE or REAL, followed by your explanation. 
ASSISTANT:
\end{lstlisting}

\subsubsection{Processing Configuration}
\begin{itemize}\setlength{\parskip}{0pt}\setlength{\itemsep}{0pt}
\item Maximum Generated Tokens: 100
\item Image Format: RGB (converted if necessary)
\item Batch Size: 1 (sequential processing)
\item Padding: Enabled
\item Special Token Handling: Customized EOS and PAD tokens
\end{itemize}

\subsubsection{Evaluation Setup}
\begin{itemize}\setlength{\parskip}{0pt}\setlength{\itemsep}{0pt}
\item Dataset Size: 10,000 samples
\item Output Processing: First word extraction (FAKE/REAL)
\item Label Mapping: FAKE → 0, REAL → 1
\item Default Fallback: FAKE (conservative approach)
\item Progress Monitoring: Accuracy reported every 100 samples
\end{itemize}

\subsubsection{Hardware Requirements}
\begin{itemize}\setlength{\parskip}{0pt}\setlength{\itemsep}{0pt}
\item GPU: NVIDIA RTX 4090
\item VRAM Usage: ~16GB
\item Storage: 30GB for model and dataset
\item Processing Time: Approximately 2 hours for 10k samples
\end{itemize}

Results from this configuration achieved 59.34\% accuracy on binary classification. Despite being designed for video understanding, the model demonstrated moderate capability in static image analysis for fake news detection.

\subsection{InstructBLIP Testing Setup}
\label{appendix:instructblip}

\subsubsection{Model Configuration}
\begin{itemize}\setlength{\parskip}{0pt}\setlength{\itemsep}{0pt}
\item Base Model: \texttt{Salesforce/instructblip-vicuna-7b}
\item Model Size: 7B parameters
\item Model Type: Conditional Generation with BLIP architecture
\item Quantization: 4-bit quantization with NF4 format
\item Double Quantization: Enabled
\item Dtype: BFloat16
\item Device Mapping: Automatic
\end{itemize}

\subsubsection{Prompt Template}
\begin{lstlisting}
Analyze this image and its title: "{title}"

Is this a fake news post or a real news post? 
Respond with either FAKE or REAL, followed by your explanation.
\end{lstlisting}

\subsubsection{Generation Parameters}
\begin{itemize}\setlength{\parskip}{0pt}\setlength{\itemsep}{0pt}
\item Maximum Length: 256 tokens
\item Minimum Length: 1 token
\item Number of Beams: 5
\item Top-p: 0.9
\item Repetition Penalty: 1.5
\item Length Penalty: 1.0
\item Temperature: 1.0
\item Sampling: Disabled
\end{itemize}

\subsubsection{Processing Configuration}
\begin{itemize}\setlength{\parskip}{0pt}\setlength{\itemsep}{0pt}
\item Image Format: RGB (converted if necessary)
\item Batch Size: 1 (sequential processing)
\item Memory Optimization: 4-bit quantization
\item Special Token Handling: Skip special tokens in decoding
\end{itemize}

\subsubsection{Evaluation Setup}
\begin{itemize}\setlength{\parskip}{0pt}\setlength{\itemsep}{0pt}
\item Dataset Size: 55,000 samples
\item Output Processing: First word extraction (FAKE/REAL)
\item Label Mapping: FAKE → 0, REAL → 1
\item Default Fallback: FAKE (conservative approach)
\item Progress Monitoring: Accuracy reported every 100 samples
\end{itemize}

\subsubsection{Hardware Requirements}
\begin{itemize}\setlength{\parskip}{0pt}\setlength{\itemsep}{0pt}
\item GPU: NVIDIA RTX 4090
\item VRAM Usage: ~10GB (with quantization)
\item Storage: 25GB for model and dataset
\item Processing Time: Approximately 5 hours for full dataset
\end{itemize}

Results from this configuration achieved 59.35\% accuracy on binary classification. The model demonstrates performance comparable to Video-LLaVA despite their architectural differences, suggesting similar capabilities in fake news detection tasks.

\end{document}